\documentclass[10pt,twocolumn,letterpaper]{article}

\usepackage{cvpr}              %

\usepackage[dvipsnames]{xcolor}

\definecolor{cvprblue}{rgb}{0.21,0.49,0.74}
\definecolor{lightgreen}{cmyk}{0.19, 0.09, 0.18, 0}
\definecolor{orange}{cmyk}{0.0391, 0.0781, 0.117, 0}

\usepackage{booktabs}
\usepackage{colortbl}
\usepackage[pagebackref,breaklinks,colorlinks,citecolor=cvprblue]{hyperref}
\usepackage{multirow}
\usepackage{makecell}
\usepackage[font=small]{caption}
\usepackage{arydshln}
\usepackage{algorithm}
\usepackage{algpseudocode}
\usepackage{listings}
\usepackage{pgfplots}
\usepackage{setspace}

\pgfplotsset{compat=1.17}

\newcommand{\app}{\raise.17ex\hbox{$\scriptstyle\sim$}}

\newcolumntype{x}[1]{>{\centering\arraybackslash}p{#1pt}}
\newcolumntype{y}[1]{>{\raggedright\arraybackslash}p{#1pt}}
\newcolumntype{z}[1]{>{\raggedleft\arraybackslash}p{#1pt}}
\newlength\savewidth

\newcommand{\sugarcrepe}{\textsc{SugarCrepe}\xspace}

\usepackage{etoolbox}
\makeatletter
\AfterEndEnvironment{algorithm}{\let\@algcomment\relax}
\AtEndEnvironment{algorithm}{\kern2pt\hrule\relax\vskip3pt\@algcomment}
\let\@algcomment\relax
\newcommand\algcomment[1]{\def\@algcomment{\footnotesize#1}}
\renewcommand\fs@ruled{\def\@fs@cfont{\bfseries}\let\@fs@capt\floatc@ruled
  \def\@fs@pre{\hrule height.8pt depth0pt \kern2pt}%
  \def\@fs@post{}%
  \def\@fs@mid{\kern2pt\hrule\kern2pt}%
  \let\@fs@iftopcapt\iftrue}
\makeatother

\usepackage[numbers]{natbib}
\def\paperID{7532} %
\def\confName{CVPR}
\def\confYear{2024}

\title{Efficient Vision-Language Pre-training by Cluster Masking}

\author{Zihao Wei\thanks{Equal contribution. Author order was determined by a coin flip.} \qquad Zixuan Pan\footnotemark[1] \qquad Andrew Owens\\
University of Michigan\\
}

\begin{document}
\maketitle
\begin{abstract}

We propose a simple strategy for masking image patches during visual-language contrastive learning that improves the quality of the learned representations and the training speed. 
During each iteration of training, we randomly mask  clusters of visually similar image patches, as measured by their raw pixel intensities. This provides an extra learning signal, beyond the contrastive training itself, since it forces a model to predict words for masked visual structures solely from context. It also speeds up training by reducing the amount of data used in each image. We evaluate the effectiveness of our model by pre-training on a number of benchmarks, finding that it outperforms other masking strategies, such as FLIP, on the quality of the learned representation.

\end{abstract}

\section{Introduction}

Images contain a great deal of redundant information, making it challenging to efficiently learn  representations from them at scale. Recent work has addressed this problem by masking image patches during vision-language contrastive learning~\cite{flip,li2022semmae,yang2022attentive,dong2023maskclip}. One simple approach is to drop a large fraction of the patches at random, making training more efficient by reducing the computational cost and memory usage in each training iteration~\cite{flip}. An alternative strategy %
is to mask sets of semantically related patches~\cite{li2022semmae,yang2022attentive,dong2023maskclip}, such as those that belong to the same object. This forces the learned model to predict words that describes missing scene structures from context, improving the learned representation. However, this approach requires a separate mechanism to group together semantically related patches, which adds considerable complexity to the learning procedure and is computationally expensive.

\begin{figure}[t]
  \centering
  \begingroup
  \setlength{\tabcolsep}{2pt}
  \resizebox{1.0\linewidth}{!}{

  \begin{tabular}{cccc}
    \vspace{0.018\columnwidth}
    \hspace{-3pt} \rotatebox[origin=c]{90}{\small Original Image}
    &
    \begin{minipage}[b]{0.33\columnwidth}
		\centering
		\raisebox{-.45\height}{\includegraphics[width=\linewidth]{fig/teaser/teaser_original_1.pdf}}
	\end{minipage}
    & 
    \begin{minipage}[b]{0.33\columnwidth}
		\centering
		\raisebox{-.45\height}{\includegraphics[width=\linewidth]{fig/teaser/teaser_original_2.pdf}}
	\end{minipage}

   &    \begin{minipage}[b]{0.33\columnwidth}
		\centering
		\raisebox{-.45\height}{\includegraphics[width=\linewidth]{fig/teaser/teaser_original_3.pdf}}
	\end{minipage}
    
    \\

    \vspace{0.018\columnwidth}
    
    \hspace{-3pt}  \rotatebox[origin=c]{90}{\small Random Mask}
    &
        \begin{minipage}[b]{0.33\columnwidth}
		\centering
		\raisebox{-.45\height}{\includegraphics[width=\linewidth]{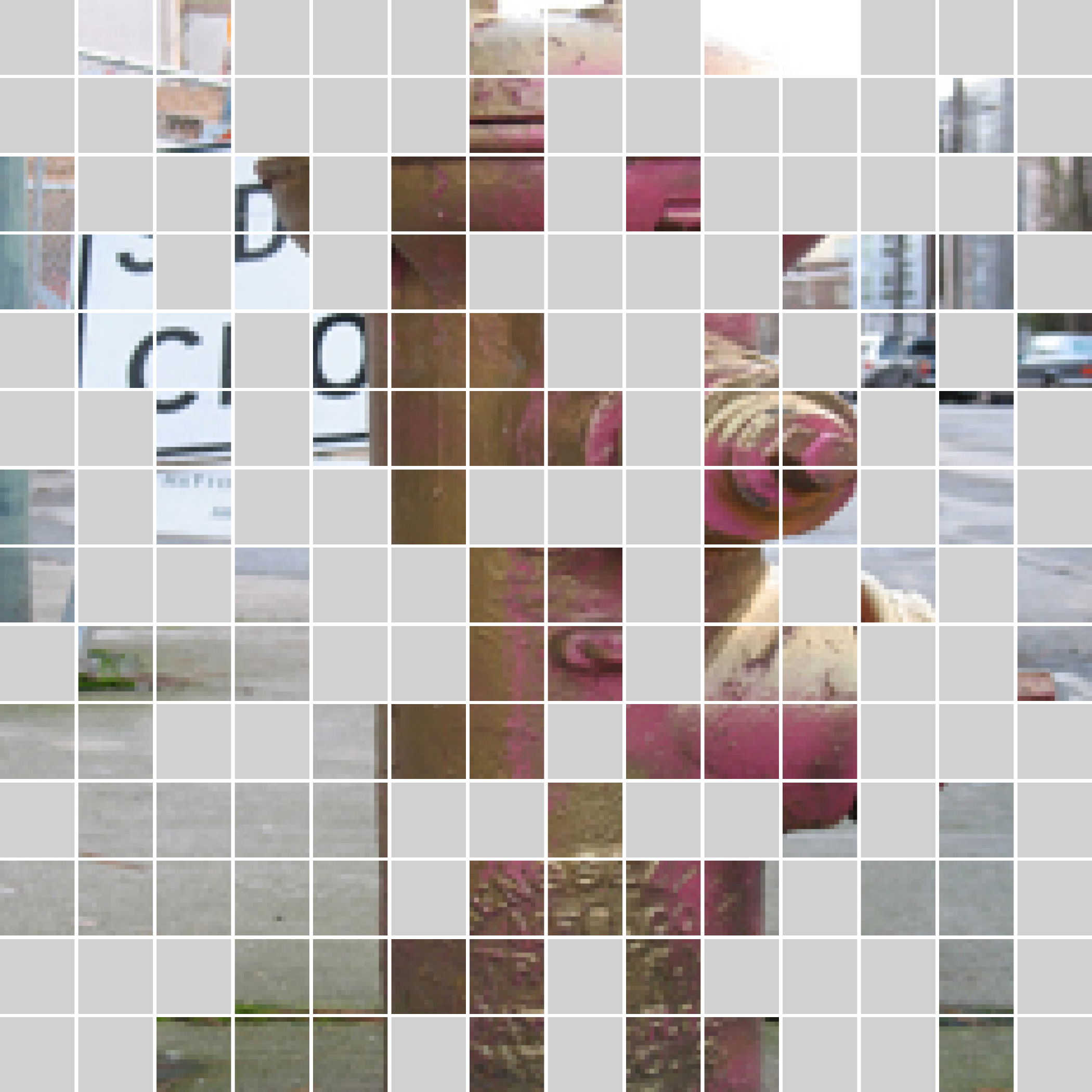}}
	\end{minipage}
    &

    \begin{minipage}[b]{0.33\columnwidth}
		\centering
		\raisebox{-.45\height}{\includegraphics[width=\linewidth]{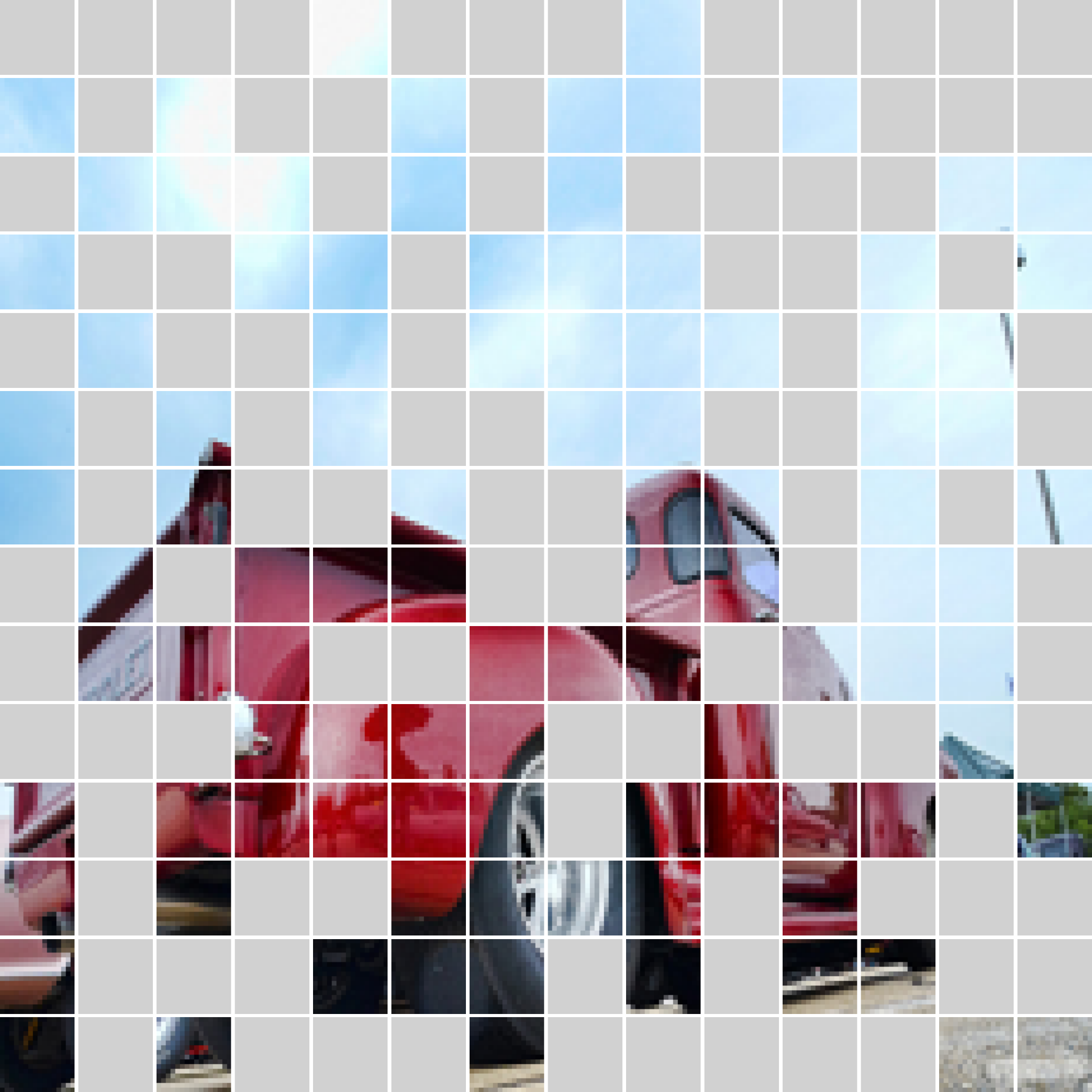}}
	\end{minipage}

    &
            \begin{minipage}[b]{0.33\columnwidth}
		\centering
		\raisebox{-.45\height}{\includegraphics[width=\linewidth]{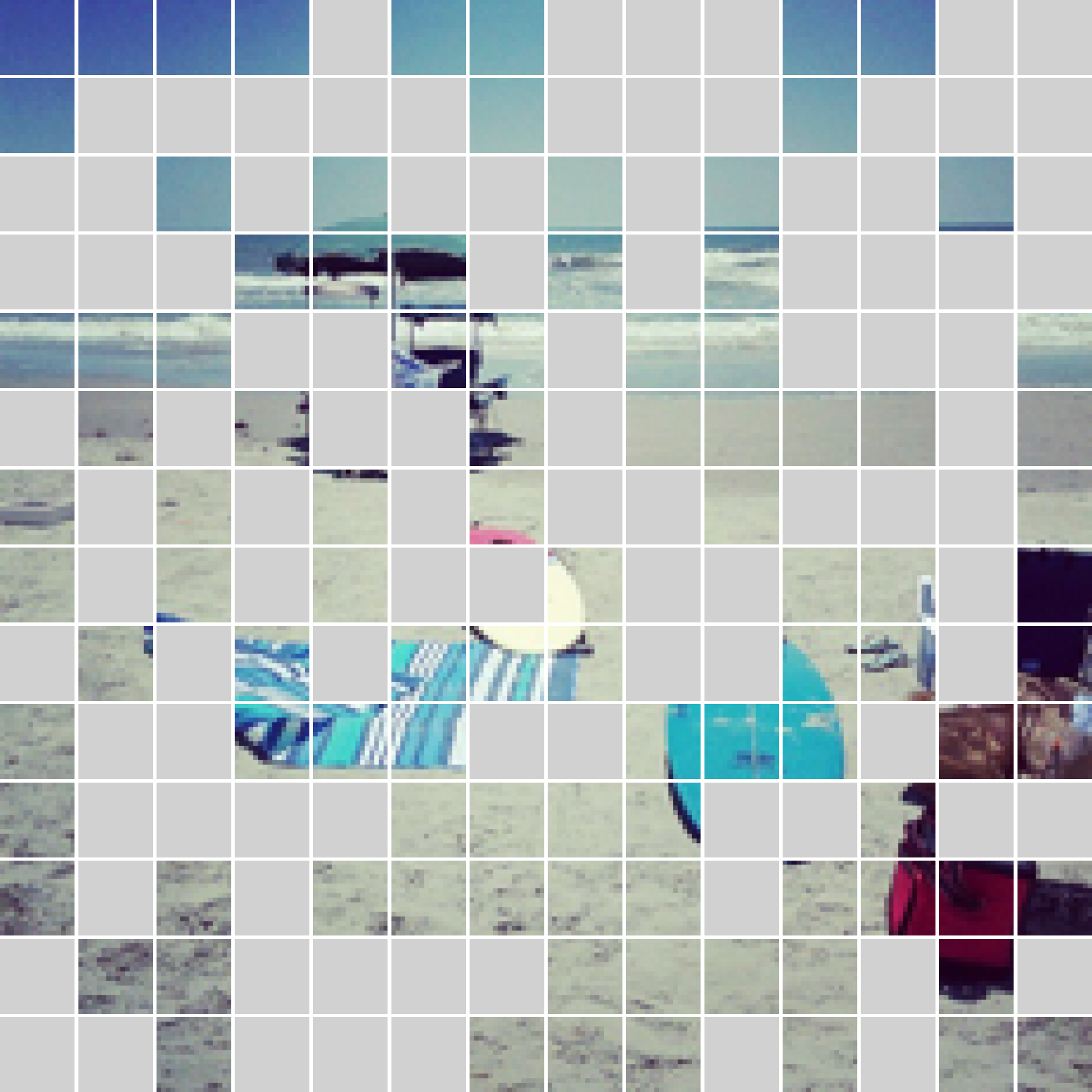}}
	\end{minipage} 

    \\

    \vspace{0.018\columnwidth}

    \hspace{-3pt}  \rotatebox[origin=c]{90}{\small Cluster Masking}  
    &
        \begin{minipage}[b]{0.33\columnwidth}
		\centering
		\raisebox{-.45\height}{\includegraphics[width=\linewidth]{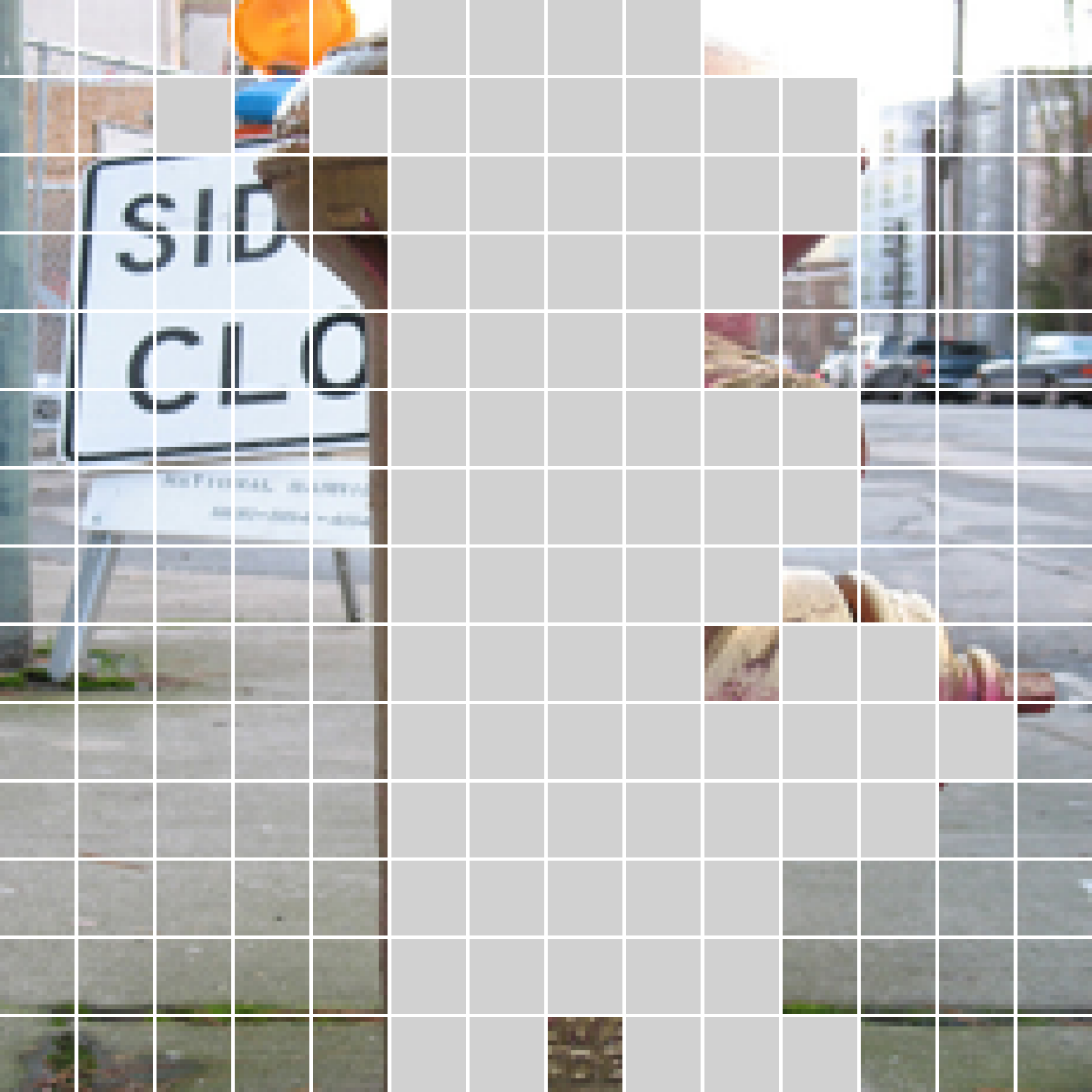}}
	\end{minipage}
    & 
    \begin{minipage}[b]{0.33\columnwidth}
		\centering
		\raisebox{-.45\height}{\includegraphics[width=\linewidth]{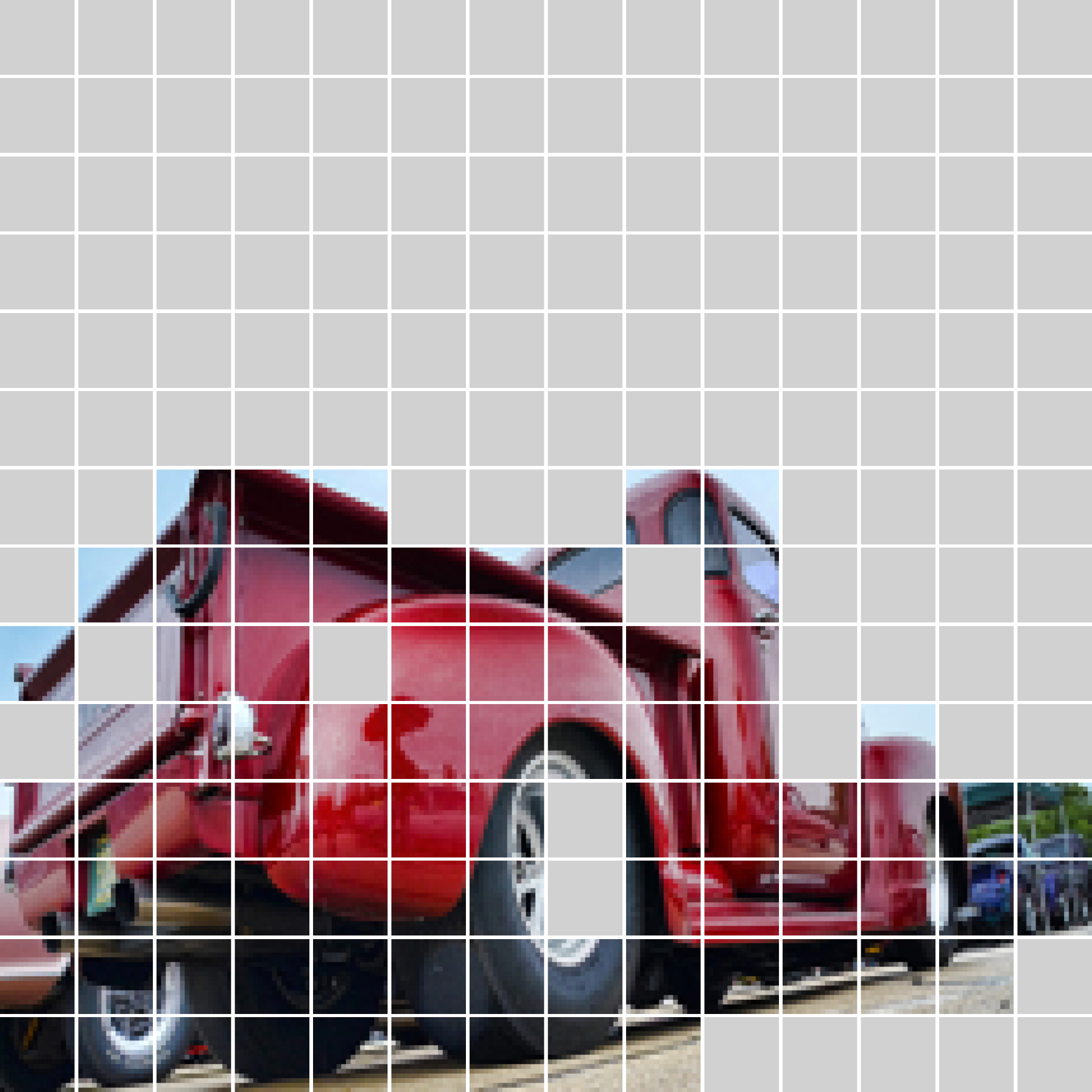}}
	\end{minipage}

    &  \begin{minipage}[b]{0.33\columnwidth}
		\centering
		\raisebox{-.45\height}{\includegraphics[width=\linewidth]{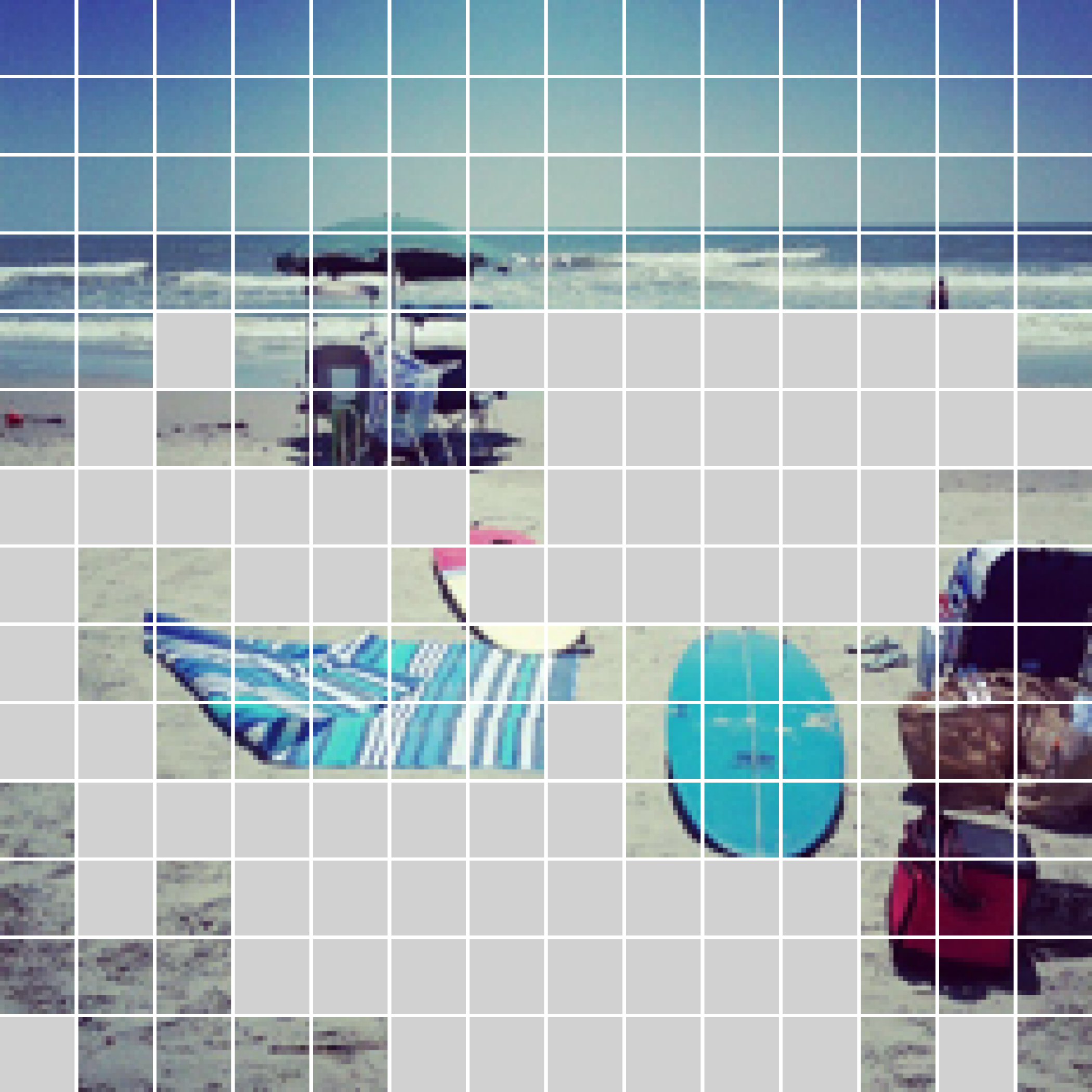}}
	\end{minipage}
    \\
   
    \rotatebox[origin=c]{90}{\small Caption}
    &
    \makecell{\small \textbf{(a)}  A red and gold \\ \small painted fire \\ \small hydrant on the street.}
    &\makecell{\small \textbf{(b)} A burgundy low-  \\ \small rider show-quality, \\ \small hot rod Chevy truck.} 
    &\makecell{\small \textbf{(c)} Two surf boars \\ on a beach \\ near the water.} \hspace{-11pt}   
    \\
  \end{tabular}
  }
  \endgroup
  \vspace{-2mm}
        \caption {{\bf Cluster masking.} We mask random {clusters} of visually similar image patches when training contrastive vision-language models (bottom). This masking strategy distinguishes our approach from methods that independently mask image patches for efficiency~\cite{flip} (middle), while providing a similar improvement in training speed. It provides an extra learning signal, since it forces a model to predict words for missing scene structures solely from context.}
\label{fig:teaser}
\vspace{-5mm}

\end{figure}

We propose a simple masking strategy for multimodal contrastive learning that avoids these shortcomings.
During training, we mask random {\em clusters} of patches (Fig.~\ref{fig:teaser}). For this clustering, we use the patches' raw RGB values as the feature representation.
Our approach takes advantage of the fact that simple measures of visual similarity can often capture coherent visual structures, such as object parts~\cite{shi2000normalized,felzenszwalb2004efficient}, especially when clusters are sampled randomly (Fig.~\ref{fig:teaser}). Our approach thus leads to more efficient training, like approaches that independently drop patches~\cite{flip}, while improving the learned representation via context prediction.  %

We take inspiration from masked region classification, a pre-training task widely used in vision-language models \cite{su2019vl,tan2019lxmert,chen2020uniter}. These models extract object features, then predict object labels for the randomly masked out regions. Our masking approach provides a similar training signal, since meaningful labels are included in the image caption. For example, as shown in Figure \ref{fig:teaser}(a), the model is tasked with associating the words ``fire hydrant'' with an image even with the hydrant itself is mostly masked out. %

We train our model on Conceptual 12M dataset \cite{cc12m} and evaluate our learned representation on a number of downstream tasks. These tasks include the zero-shot classification and linear probing on ImageNet~\cite{imagenet}, text and image retrieval on MS-COCO~\cite{lin2014microsoft}, and the \sugarcrepe language composition benchmark~\cite{hsieh2023sugarcrepe}. In our experiments, our model outperforms FLIP~\cite{flip} and CLIP~\cite{clip} on downstream performance, while having efficiency comparable with FLIP. We also show that the performance can further be improved by using the model's learned feature embedding during clustering.

\section{Related Work}
\paragraph{Contrastive Vision-Language Pre-training.}

Vision-Language Pre-training (VLP) focuses on establishing connections between images or their components and human-interpretable language. This field initially evolved from transferring supervised learning models, which incorporated object detection modules to generate fine-grained visual labels \cite{su2019vl, tan2019lxmert, chen2020uniter}. Subsequently, there was a shift towards large-scale learning using noisy web data, moving away from reliance on fine-grained labels \cite{radford2021learning, jia2021scaling, yu2022coca, alayrac2022flamingo, li2021align, yuan2021florence,yang2024binding}. A significant development in this domain was CLIP \cite{radford2021learning}, which applied contrastive learning techniques \cite{chen2020simple, he2020momentum} to train models to associate correct image-text pairs and dissociate incorrect ones. CLIP scaled contrastive visual-language models significantly beyond previous work, enabling strong feature learning and zero-shot performance.
However, further scaling significantly increases the pre-training demands, requiring larger datasets and batch sizes.

In response to these challenges, recent research has explored incorporating masking into images to reduce training time and allow for more samples per batch \cite{flip, dong2023maskclip, fu2021violet, yang2022attentive}. Methods such as MaskClip \cite{dong2023maskclip}, FLIP \cite{flip}, and VIOLET \cite{fu2021violet} have implemented random masking strategies. Yet, it has been noted that random masking may not be as effective on relatively small datasets \cite{yang2022attentive, li2023clipav2}. To address this, ACLIP \cite{yang2022attentive} introduced a method of masking tokens with low cross-attention scores with text. However, this approach necessitates two forward passes to generate the attention map and requires additional computational modules \cite{mu2021slip}. In our work, we aim to avoid these limitations, proposing an effective masking method that is based on a patch's raw RGB values.

\paragraph{Masked Image Modeling.}
In the field of language modeling, the effectiveness of models that learn to reconstruct corrupted inputs for generating robust features has been recognized \cite{kenton2019bert,liu2019roberta}. This approach, known as Mask Language Modeling (MLM), has been adapted in the realm of image processing as Mask Image Modeling (MIM). MIM techniques involve reconstructing either image patches or their features \cite{bao2021beit, dong2023peco, xie2022simmim, wei2022masked,chen2020generative,he2021masked,chen2022context,wang2023SwinMM,wei2023mae}. The pioneering work in BEIT \cite{bao2021beit} introduced the reconstruction of discrete tokens, akin to VQ-VAE \cite{van2017neural}, using block-wise masking. This method demonstrated results on par with contrastive learning and self-distillation methods \cite{caron2021emerging, mocov3} during model fine-tuning. Later approaches include PeCo's \cite{dong2023peco} novel visual codebook learning method and BEIT V2's \cite{peng2022beit} integration of self-distillation methods, using a teacher-student backbone and feature-level KL divergence loss \cite{touvron2021training}. Further exploration in this field has led to the use of natural image signals as reconstruction targets, moving away from learned features. Examples include SimMIM \cite{xie2022simmim}, which reconstructs pure RGB values, MaskFeat \cite{wei2022masked}, introducing reconstruction of the Histogram of Oriented Gradients (HOG) features, and MAE \cite{he2021masked}, which reconstructs pixel-normalized RGB values. Our work draws inspiration from these studies, particularly in using pixel-normalized RGB values to compute patch similarities, arguing for a more effective distribution of patch features.

\paragraph{Masking Strategies in MIM.}
Parallel investigations have focused on masking strategies in MIMs \cite{li2021mst, kakogeorgiou2022attmask, li2022semmae, wilf2023difference,shi2022adversarial,feng2023evolved,xie2022masked}. Early works like BEIT and its successors used block masking, while others such as SimMIM, MaskFeat, and MAE applied random patch-wise masking. Attention-based masking strategies have also been explored, typically using attention maps from vision transformers. MST \cite{li2021mst} masks less essential parts with low attention scores, using a reconstruction loss approach. In contrast, AttnMask \cite{kakogeorgiou2022attmask} masks highly attentive patches and applies self-distillation loss. These methods involve simultaneous updates of attention maps and masks during training. A potential limitation of this approach is that insufficiently trained attention maps may not capture structured features effectively. SemMAE \cite{li2022semmae}, starting with iBot features \cite{zhou2021ibot}, adopts an easy-to-hard masking strategy, starting with masking parts within clusters and gradually expanding to entire clusters. Wilf et al. \cite{wilf2023difference} introduce a unique entity-reinforced language model for masking objects in video frames. However, the reliance on pre-trained features or extracting attention maps can be computationally intensive. Evolved Part Masking \cite{feng2023evolved} proposes using EM algorithm on attention maps to get a clustering before performing SemMAE style masking. Our approach also adopts a cluster based masking strategy in vision-language pre-training, enabling faster pre-training without requiring additional modifications to the model.

\section{Method}
We propose a cluster based masking strategy for contrastive vision-language pre-training, focusing on masking random clusters with visually similar semantics. Our method selects random anchor patches as cluster centers and computes pairwise patch distances to form clusters. These clusters are then masked entirely. To enhance accuracy in cluster formation, we introduce an adaptive layer for refining the distance matrix. Additionally, attention masks and a hard patch cutoff are used to ensure uniform input sizes are consistent in batches for auto differentiation.

\subsection{Contrastive Vision-Language Pre-training}

Our approach builds on contrastive vision-language pre-training methods, such as CLIP \cite{clip}. We use contrastive learning to align embeddings of matching text-image pairs and separate those of non-matching pairs. This process is steered by two symmetric InfoNCE losses \cite{oord2018representation}: the vision-to-language loss $\mathcal{L}_{v \rightarrow l}$ and its counterpart, the language-to-vision loss $\mathcal{L}_{l \rightarrow v}$. The vision-to-language loss is defined as:
{\small
\begin{equation}
\setlength{\abovedisplayskip}{3pt}
\setlength{\belowdisplayskip}{3pt}
\mathcal{L}_{v \rightarrow l} = - \log \frac{\exp(\text{sim}(I, T) / \tau)}{\sum_{j=1}^{N} \exp(\text{sim}(I, T_j) / \tau)},
\end{equation}}%
where $I$ and $T$ are the embeddings for the image and text respectively, $\text{sim}$ denotes the similarity function (we use a dot product), and $\tau$ is a temperature parameter. Similarly, $\mathcal{L}_{l \rightarrow v}$ is formulated by normalizing the loss using a batch of $N$ image examples $\{I_j\}_{i=1}^N$ instead of text examples $\{T_j\}_{i=1}^N$.

\begin{figure}[t]
  \centering
  \setlength{\tabcolsep}{2pt}{
  
  \begin{tabular}{ccc}
        \begin{minipage}[b]{0.3\columnwidth}
		\centering
		{\includegraphics[width=\linewidth]{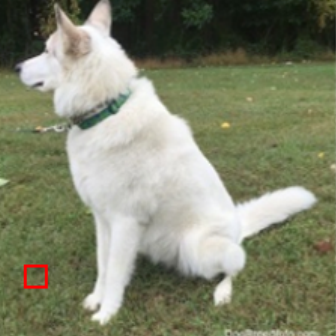}}
	      \end{minipage}
    & 
        \begin{minipage}[b]{0.3\columnwidth}
		\centering
		{\includegraphics[width=\linewidth]{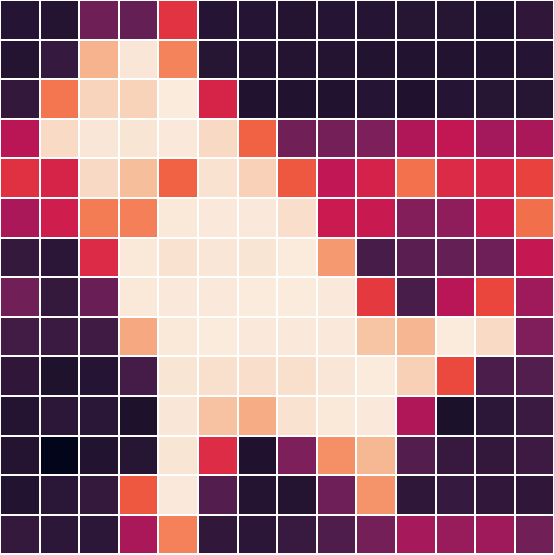}}
	      \end{minipage}
    &
        \begin{minipage}[b]{0.3\columnwidth}
		\centering
		{\includegraphics[width=\linewidth]{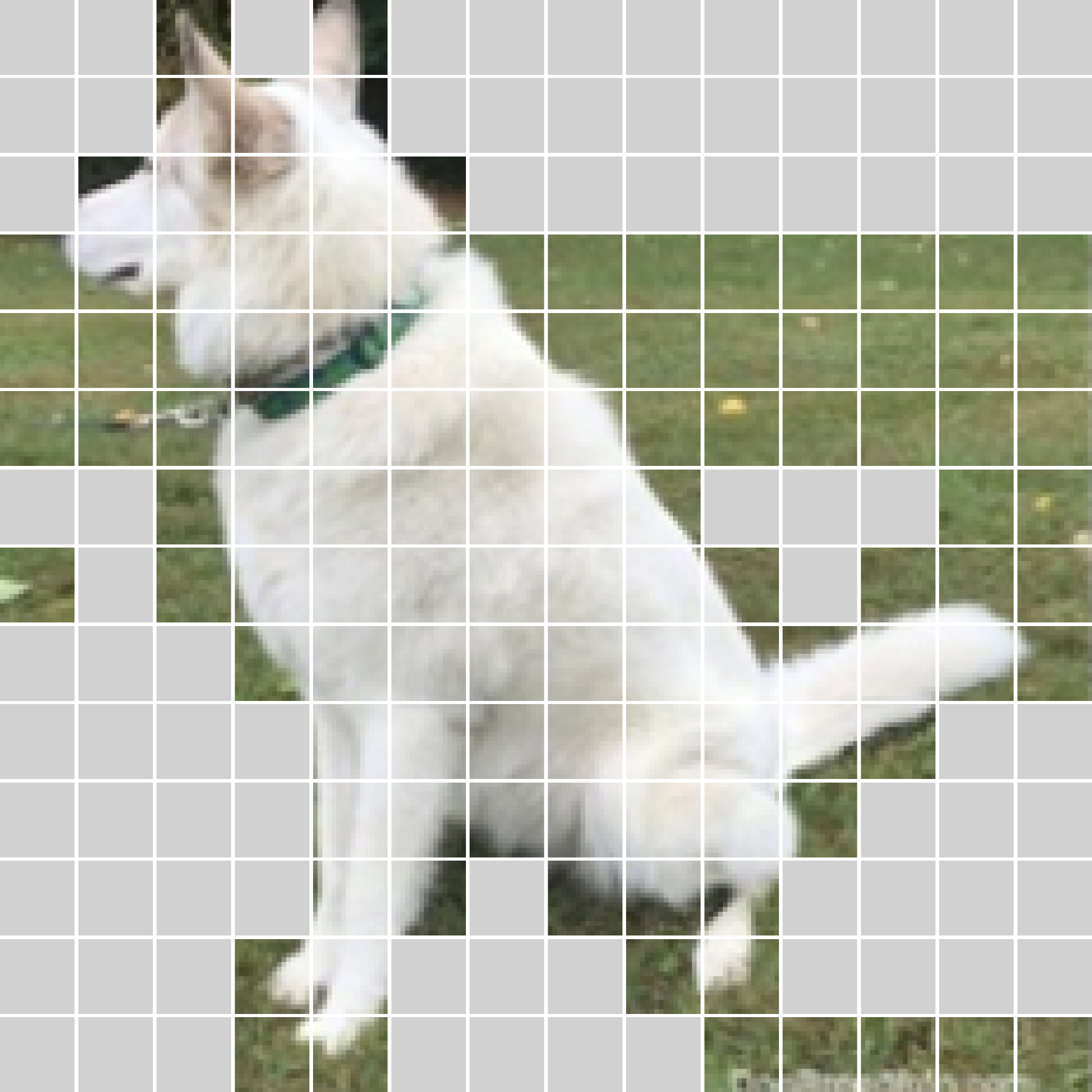}}
	    \end{minipage}
    \\
    \small{Anchor Patch} & \small{Distance Heatmap} & \small{Masked Image}\\ %
  \end{tabular}
  }
  \vspace{-2mm}
  \caption {\textbf{Choosing clusters.} The process begins by randomly selecting anchor patches from the image. We then calculate the pairwise distances among all patches. Clusters formed within a distance threshold are masked out. We show cluster obtained from a single anchor patch.} 
\label{fig:model}
\vspace{-5mm}

\end{figure}

\subsection{Cluster Masking}

We introduce a masking strategy that drops out random clusters. While one option would be to use an off-the-shelf clustering method, such as K-Means \cite{lloyd1982least}, we choose instead to use a simple and efficient method that results in a random clustering each training iteration (Figure \ref{fig:model}). Our approach resembles a single iteration of K-Means, and works by selecting a set of exemplar patches, which each define a cluster. In our experiments, we also evaluate masking clusters obtained using K-Means as an alternative approach. %

We split an input $H \times W$ image into patches, following \cite{dosovitskiy2020vit}. We then compute the pairwise cosine similarity between every pair of normalized patches, which we use as a distance function $d(\mathbf{x},\mathbf{y})$. We choose a small subset (less than 5\%) of these patches at random to act as cluster centers. For each of these selected {\em anchor patches}, we define a cluster consisting of patches that lie within a distance $r$.  The cluster for an exemplar patch ${\mathbf x}$ is represented by: $\mathrm{S}_{\mathbf x} = \{ \mathbf{y} \mid d(\mathbf{x},\mathbf{y}) \leq r\}$ for image patches. %

All patches within a cluster are masked out. The distance threshold $r$ is automatically searched before training according to an average masking ratio. We provide a simplified pseudocode of the masking strategy in Algorithm \ref{alg:mask}. 

\paragraph{Clustering Embedding Features.}
Another variant of patch feature is the combination of pure RGB values and patch embedding layer features from transformers \cite{dosovitskiy2020vit}. When computing similarity scores, we integrate these two measures into a weighted sum, where the weight of each measure is determined by:
{
\small
\begin{equation}
\setlength{\abovedisplayskip}{3pt}
\setlength{\belowdisplayskip}{3pt}
d(\mathbf{x},\mathbf{y}) = \alpha \cdot d_{rgb}(\mathbf{x},\mathbf{y}) + (1- \alpha) \cdot d_{emb}(\mathbf{x},\mathbf{y}),
\end{equation}
}
where $\mathbf{x}$ and $\mathbf{y}$ represent two patches, $d_{rgb}$ is the cosine similarity based on pure RGB values, and $d_{emb}$ is the cosine similarity based on the transformer's embedding features. The weight parameter $\alpha$ linearly increases from 0 to 1 during training.

The embedding layer is calculated before the patches enter the transformer, thus we could reuse the patch embeddings in the transformer without computing them twice. Using the embedding layer is advantageous because it incorporates positional encodings \cite{Transformer}. This integration potentially introduces spatial constraints, which we believe can further enhance our masking strategy.

\paragraph{Handling Batched Inputs.}
Deep learning libraries, like PyTorch \cite{paszke2017automatic}, typically process batched inputs of uniform size. However, in our method, the mask ratio would vary across different images, leading to fluctuations in the number of patches. To accelerate the process, we introduce a \textit{minimum mask ratio} threshold $\beta$ for each image. If the calculated mask ratio for an image doesn't meet this predefined threshold, we proceed to randomly drop patches until the desired ratio is achieved. Conversely, for images with patch counts less than the threshold, we use attention masks to avoid masked parts engaging in attention calculation \cite{devlin2018bert}.

\begin{algorithm}[htbp]
\caption{Pseudocode of cluster based masking in a PyTorch-like style.}
\label{alg:code}
\algcomment{\fontsize{7.2pt}{0em}\selectfont \texttt{bmm}: batch matrix multiplication.
\vspace{-5mm}
}

\definecolor{codeblue}{rgb}{0.25,0.5,0.5}
\lstset{
  backgroundcolor=\color{white},
  basicstyle=\fontsize{7.2pt}{7.2pt}\ttfamily\selectfont,
  columns=fullflexible,
  breaklines=true,
  captionpos=b,
  commentstyle=\fontsize{7.2pt}{7.2pt}\color{codeblue},
  keywordstyle=\fontsize{7.2pt}{7.2pt},
}
\begin{lstlisting}[language=python,mathescape=true]
# img: the image for mask
# mask_ratio: the ratio for chosing anchor patches
# r: the threshold for computing cluster

$\text{\textbf{def}}$ generate_mask(img, mask_ratio, r):
    # Make image into shape (N,L,C). 
    # N: number of samples per batch 
    # L: number of patches 
    # C: size of feature dimension
    img = patchify(img)    
    # Normalize each patch
    img = (img - img.mean(dim=-1)) / img.std(dim=-1)
    
    # Compute pairwise cosine similarity matrix
    x = img / img.norm(dim=-1) 
    distance = bmm(x, x.transpose(-2, -1)) 
    
    # Generate a boolean masking matrix of shape (N, L)
    init_mask = random_patch_indicies(mask_ratio) 
    init_mask = init_mask[:, :, None].float()
    # Get the cluster-based mask
    candidates = (distance * init_mask) $\geq$ r
    cluster_mask = candidates.sum(1) $\geq$ 1
    # Mask both the anchor patches and clusters
    mask  = init_mask $\text{\textbf{or}}$ cluster_mask
    
    $\text{\textbf{return}}$ mask
\end{lstlisting}
\label{alg:mask}
\vspace{-2mm}
\end{algorithm}

\vspace{0mm}

\section{Experiments}
We present a comprehensive evaluation of our proposed algorithm to exhibit the performance, robustness, scalability, and efficiency of our framework.

\subsection{Implementation Details}

\paragraph{Datasets and Training Details.}
We train our model using the Conceptual 12M (CC12M) dataset\cite{cc12m}, containing 12 million unique image-text pairs, for pre-training our vision-language models. We use ViT-B/16 as backbone for image encoder. The text encoder is a 12-layer transformer, equipped with 8 multi-head attention units and 512-dimensional embeddings. Input images are processed at a resolution of $224 \times 224$, and text inputs are adjusted to 77 tokens, either by truncation or padding. A class token is transformed into a 512-dimensional feature embedding via a multi-layer perceptron (MLP). For optimization, we use the AdamW optimizer with a learning rate of $5 \times 10^{-4}$, $\beta_1 = 0.9$, and $\beta_2 = 0.98$. We use a batch size of 256 per GPU, and train using 8 NVIDIA A40 GPUs.

Our method comes in three variants: K-Means, RGB and Embedding. The RGB model clusters based on raw image patches, while the embedding model integrates patch embedding features with RGB for clustering. %
In the K-Means variant of the model, we mask out half of the clusters randomly. The model constructs 12 clusters and runs for a maximum of 10 iterations. For both RGB and embedding models, we set an average masking ratio of 50\%, following the recommendations of FLIP \cite{flip} for optimal masking ratios. %
In the RGB approach, we use a 50\% cutoff for $\text{Ours-RGB}_{0.5}$ and 30\% for $\text{Ours-RGB}_{0.3}$, whereas the Ours-Embedding model uses a 30\% cutoff. Also, the RGB model selects anchor patches at a 3\% ratio, compared to a 5\% ratio in the embedding model.

\paragraph{Baselines.}
In our study, we establish baselines using three models: CLIP, FLIP, and $\text{FLIP}_\text{Attn}$, each trained from scratch on the CC12M dataset. These baseline models are derived from the open-source implementation of CLIP, known as OpenClip \cite{ilharco_gabriel_2021_5143773,cherti2023reproducible, Radford2021LearningTV,schuhmann2022laionb}. For both FLIP and $\text{FLIP}_\text{Attn}$, we implement a patch dropout ratio of 50\%. Specifically, FLIP uses a random dropout approach, whereas $\text{FLIP}_\text{Attn}$ adopts an attention-based masking strategy inspired by ACLIP \cite{yang2022attentive}. This strategy involves processing the image through the encoder and then averaging across attention heads in the final transformer block to determine attention scores. Patches that receive the highest attention in relation to the [CLS] token are retained.

To ensure a fair comparison among these methods, we keep the number of patches consistent the same as ours within a single batch, which means for FLIP and $\text{FLIP}_\text{Attn}$, we apply a batch size of 256 for one GPU and for CLIP, we use 128 instead. Additionally, we apply a scaling law on learning rate across different models.

\paragraph{Evaluation Details.}
Our models are tested across various benchmarks to ensure its robustness and effectiveness. We conduct zero-shot image-to-text and text-to-image retrieval tasks on COCO \cite{lin2014microsoft} and Flickr \cite{flickr}, assessing its performance meticulously. %
Further more, we evaluate the models' image representation quality by reporting both the zero-shot classification and linear probing performance on three mainstream datasets: ImageNet \cite{imagenet}, CIFAR-10, and CIFAR-100 \cite{Krizhevsky09learningmultiple}.  %
Zero-shot results of some other datasets like ImageNet variants \cite{ImageNetA, ImageNetO, ImageNetR, ImageNetS, imagenetv2}, Caltech101 \cite{caltech101}, Flowers \cite{flower} and Pets \cite{pets}, are also reported to verify the method's robustness.

For these tasks, our approach adheres strictly to the implementation used in the CLIP benchmark, ensuring consistency and reliability in the evaluation process. Furthermore, we assess the effectiveness of our methodology on language composition tasks using \sugarcrepe \cite{hsieh2023sugarcrepe}. This evaluation aims to determine its adaptability and efficiency across various contexts, including object, attribute, and relation manipulations. Within the \sugarcrepe framework, models are tasked with identifying the correct caption that accurately describes an image, distinguishing it from a closely related but incorrect text hard negatives. The hard negatives is characterized by minor composition differences from the accurate caption.

\begin{figure*}[htbp]
  \centering
  \begingroup
  \setlength{\tabcolsep}{1pt}
  \renewcommand{\arraystretch}{2}
  \resizebox{0.8\linewidth}{!}{
  \begin{tabular}{ccccccc}
    \vspace{0.009\columnwidth}
    \rotatebox[origin=c]{90}{Images}&
    \begin{minipage}[b]{0.3\columnwidth}
		\centering
		\raisebox{-.45\height}{\includegraphics[width=\linewidth]{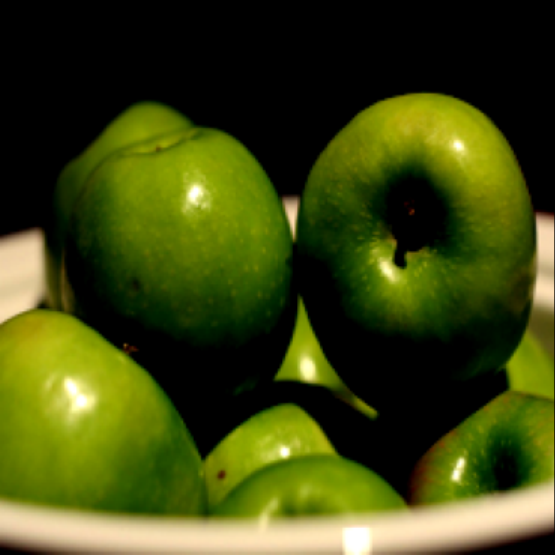}}
	\end{minipage}
    & 
    \begin{minipage}[b]{0.3\columnwidth}
		\centering
		\raisebox{-.45\height}{\includegraphics[width=\linewidth]{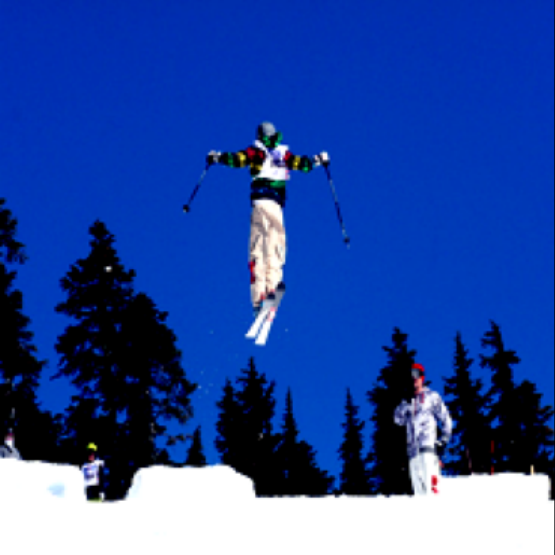}}
	\end{minipage}
    & 
    \begin{minipage}[b]{0.3\columnwidth}
		\centering
		\raisebox{-.45\height}{\includegraphics[width=\linewidth]{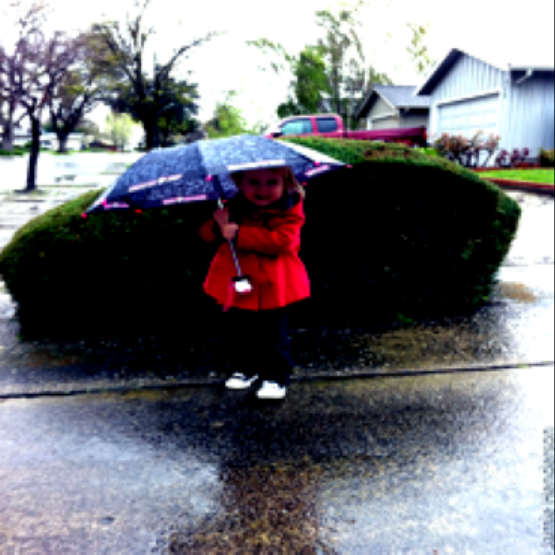}}
	\end{minipage}
    &
    \begin{minipage}[b]{0.3\columnwidth}
		\centering
		\raisebox{-.45\height}{\includegraphics[width=\linewidth]{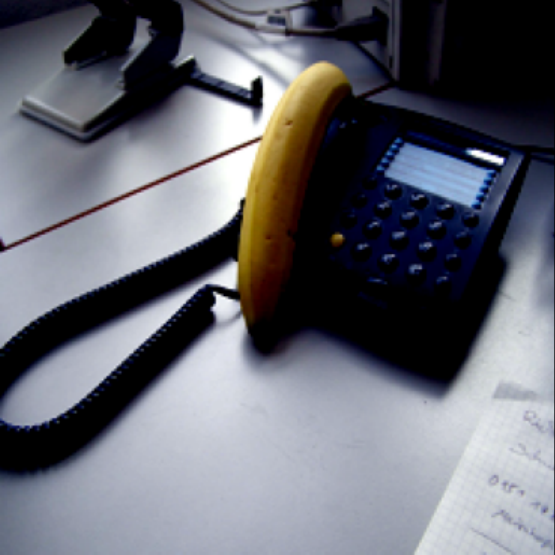}}
	\end{minipage}
    &
    \begin{minipage}[b]{0.3\columnwidth}
		\centering
		\raisebox{-.45\height}{\includegraphics[width=\linewidth]{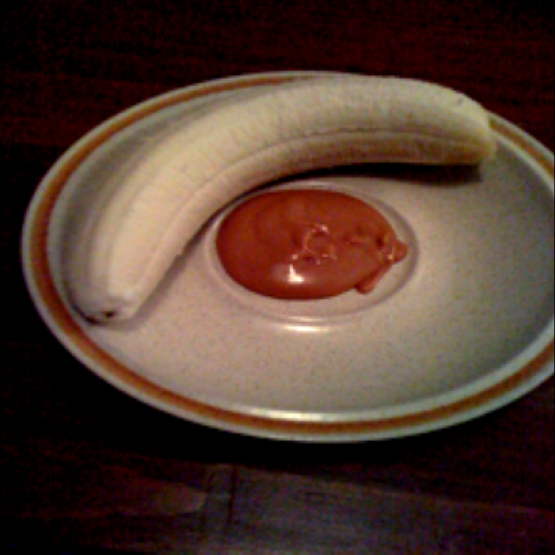}}
	\end{minipage}
    &
    \begin{minipage}[b]{0.3\columnwidth}
		\centering
		\raisebox{-.45\height}{\includegraphics[width=\linewidth]{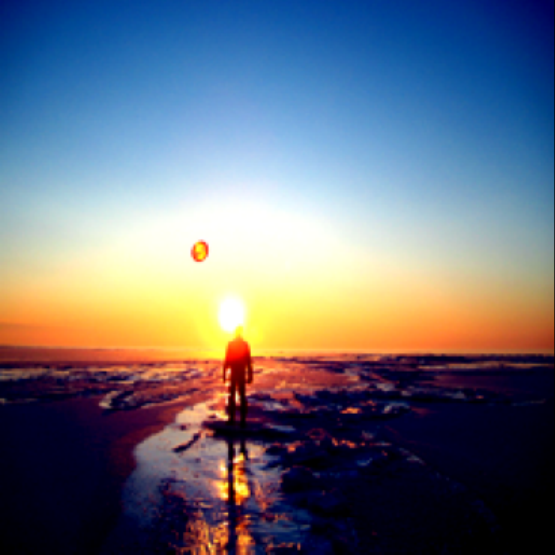}}
	\end{minipage}
    \\

    \vspace{0.009\columnwidth}
    
    \rotatebox[origin=c]{90}{Anchors}&
        \begin{minipage}[b]{0.3\columnwidth}
		\centering
		\raisebox{-.45\height}{\includegraphics[width=\linewidth]{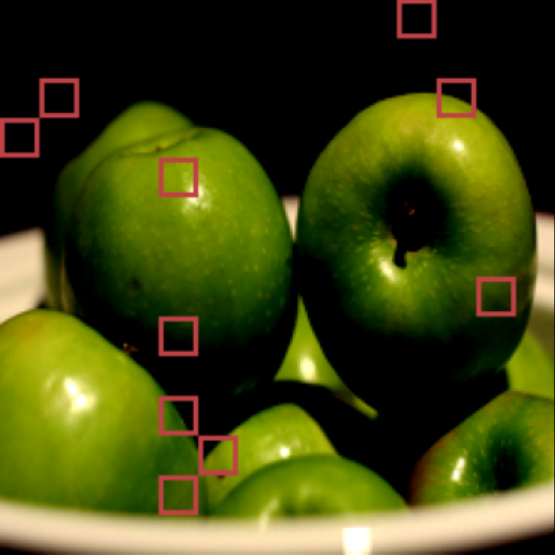}}
	\end{minipage}
    & 
    \begin{minipage}[b]{0.3\columnwidth}
		\centering
		\raisebox{-.45\height}{\includegraphics[width=\linewidth]{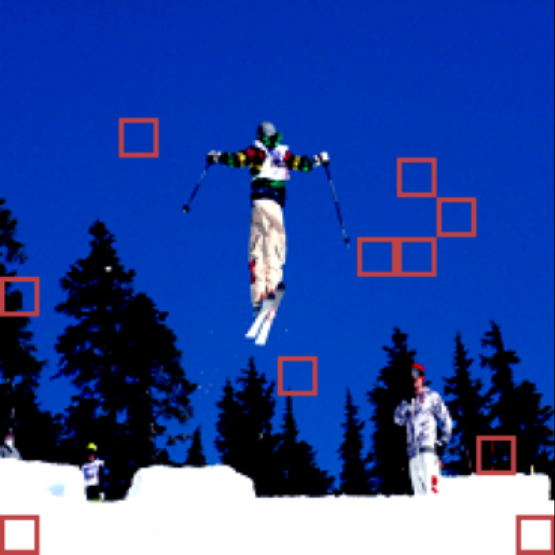}}
	\end{minipage}
    & 
    \begin{minipage}[b]{0.3\columnwidth}
		\centering
		\raisebox{-.45\height}{\includegraphics[width=\linewidth]{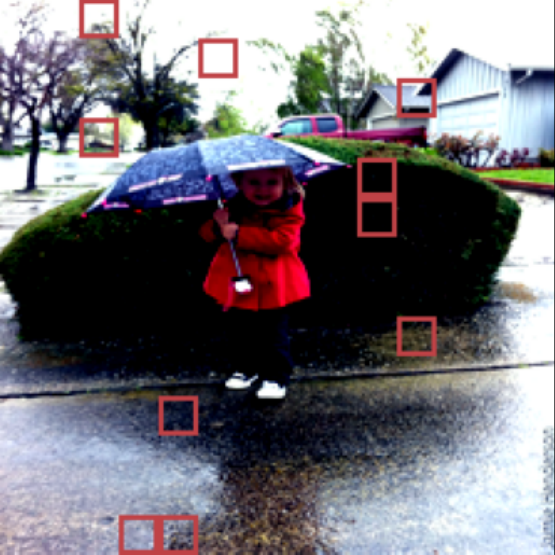}}
	\end{minipage}
    &
    \begin{minipage}[b]{0.3\columnwidth}
		\centering
		\raisebox{-.45\height}{\includegraphics[width=\linewidth]{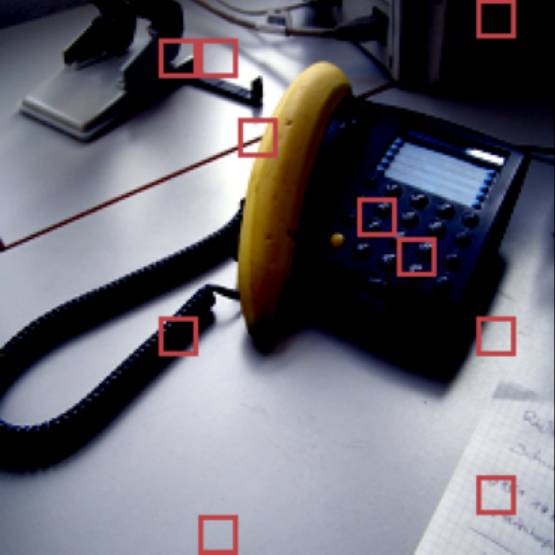}}
	\end{minipage}
    &
    \begin{minipage}[b]{0.3\columnwidth}
		\centering
		\raisebox{-.45\height}{\includegraphics[width=\linewidth]{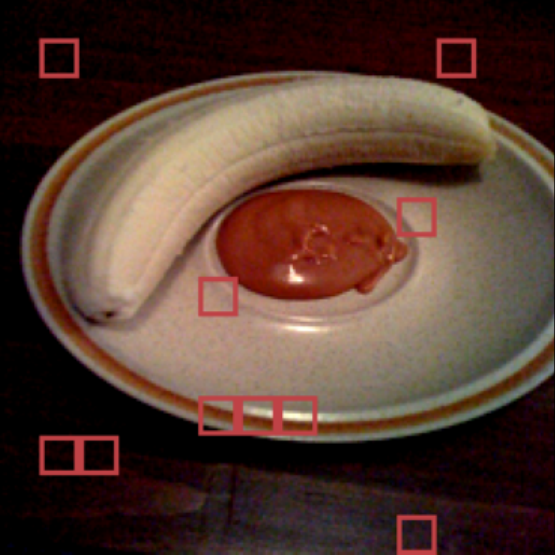}}
	\end{minipage}
    &
    \begin{minipage}[b]{0.3\columnwidth}
		\centering
		\raisebox{-.45\height}{\includegraphics[width=\linewidth]{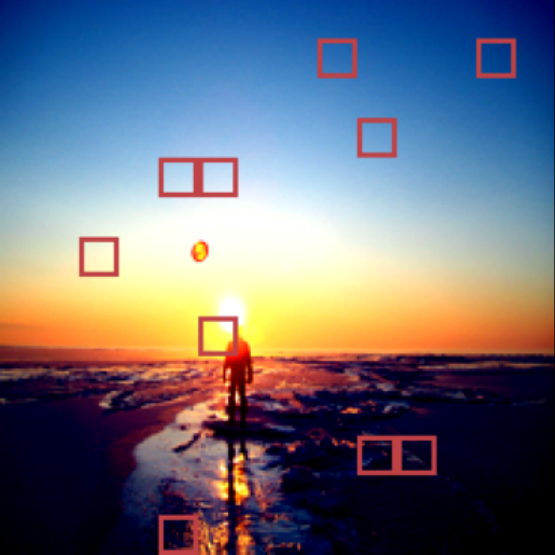}}
	\end{minipage}
    \\

    \vspace{0.009\columnwidth}

    \rotatebox[origin=c]{90}{Clusters}&
        \begin{minipage}[b]{0.3\columnwidth}
		\centering
		\raisebox{-.45\height}{\includegraphics[width=\linewidth]{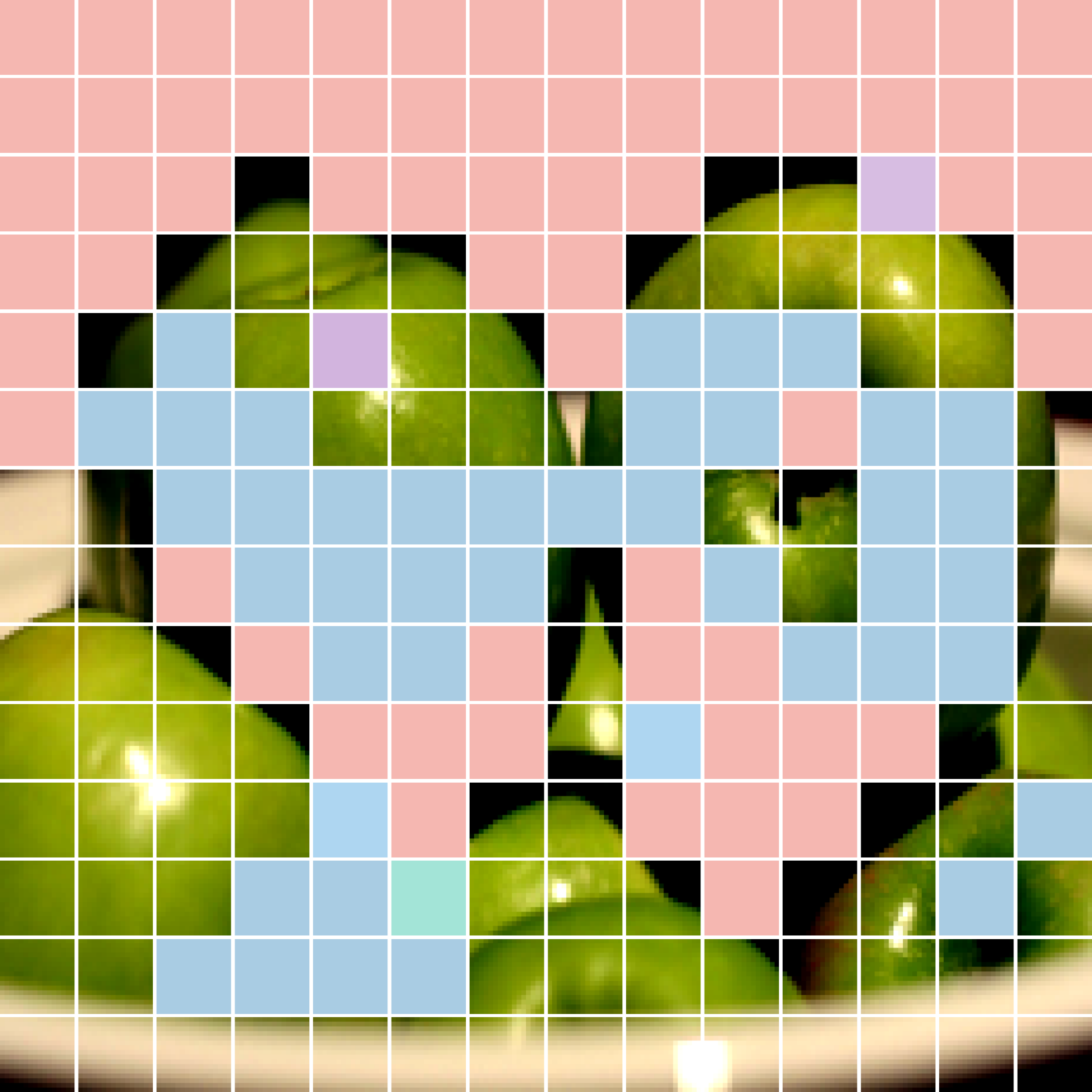}}
	\end{minipage}
    & 
    \begin{minipage}[b]{0.3\columnwidth}
		\centering
		\raisebox{-.45\height}{\includegraphics[width=\linewidth]{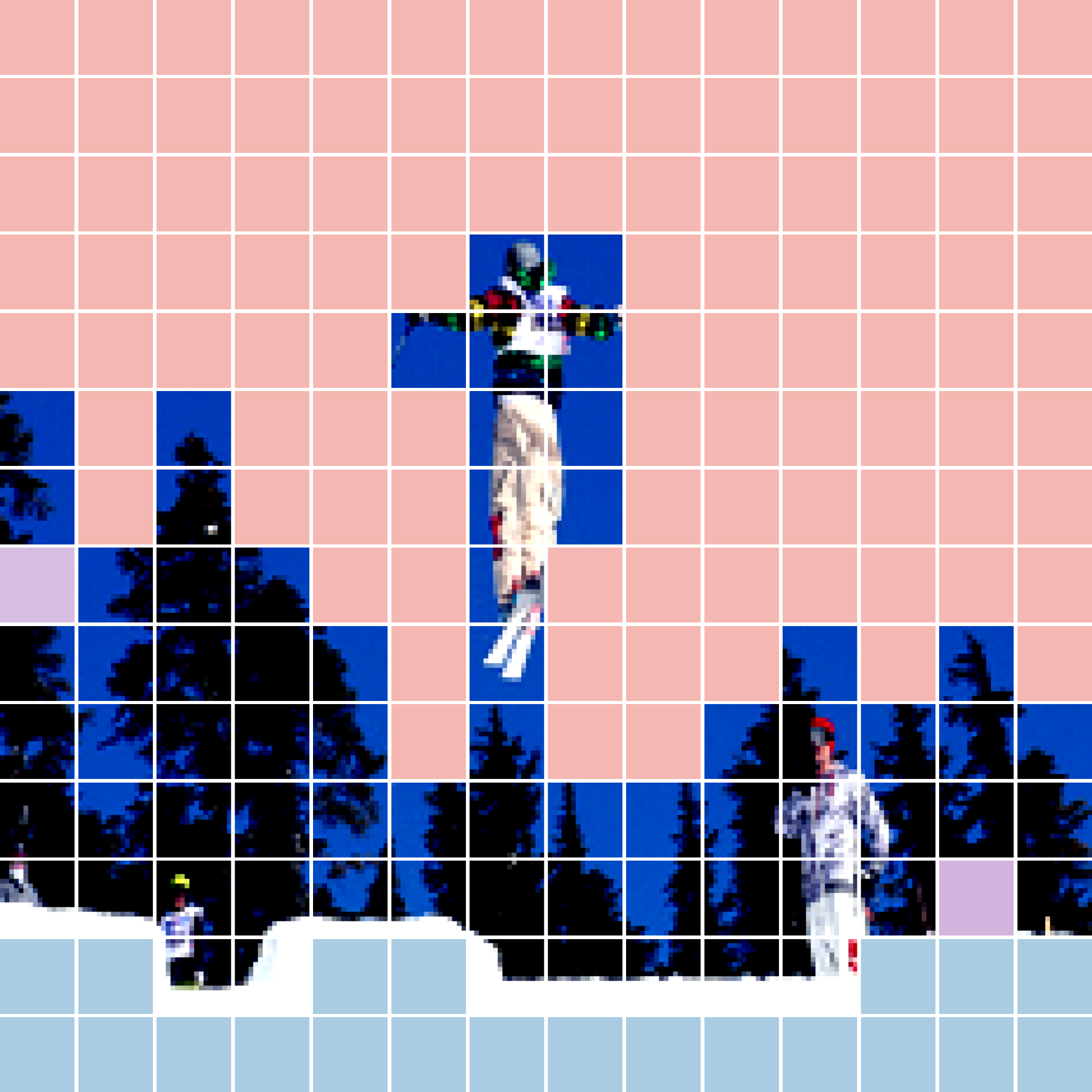}}
	\end{minipage}
    & 
    \begin{minipage}[b]{0.3\columnwidth}
		\centering
		\raisebox{-.45\height}{\includegraphics[width=\linewidth]{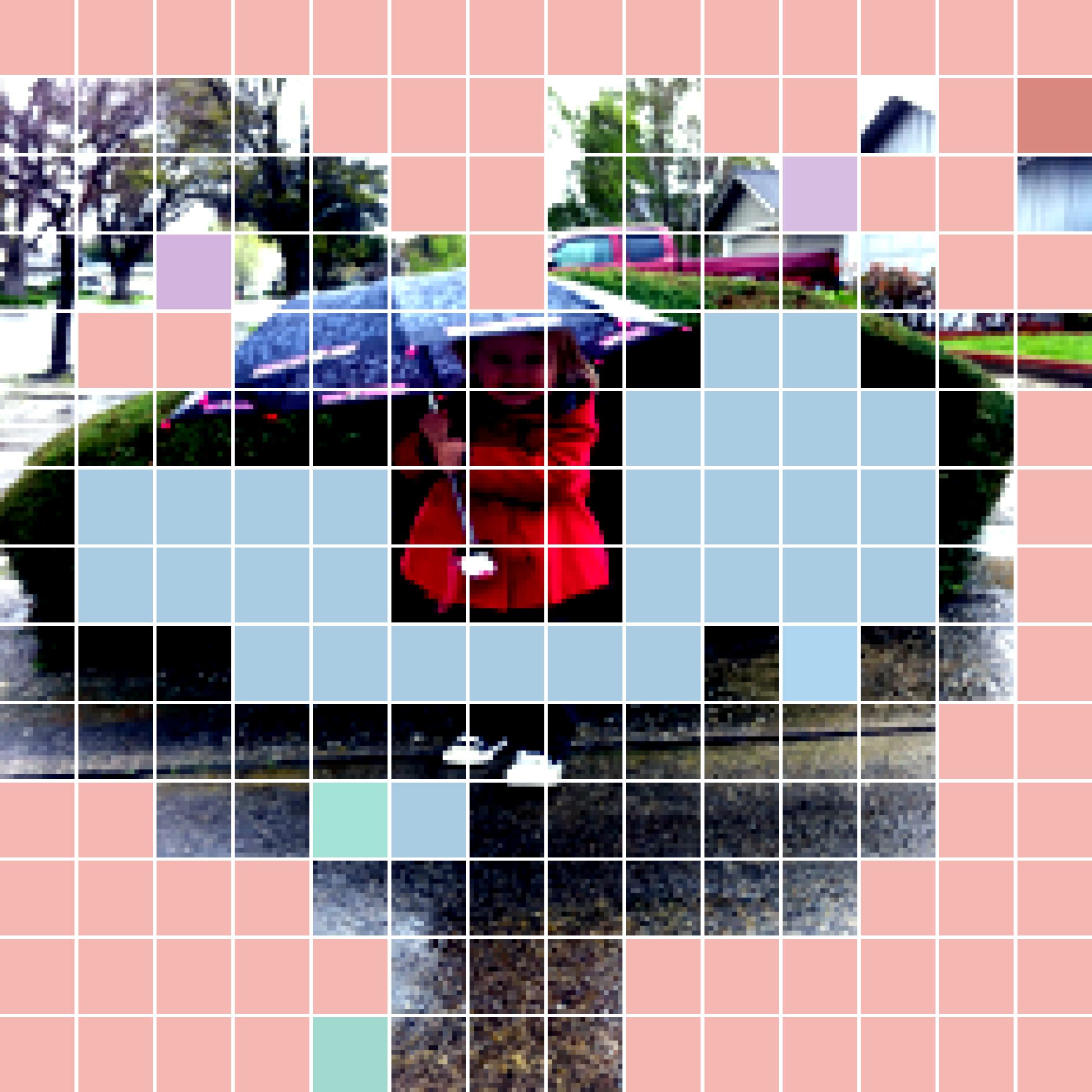}}
	\end{minipage}
    &
    \begin{minipage}[b]{0.3\columnwidth}
		\centering
		\raisebox{-.45\height}{\includegraphics[width=\linewidth]{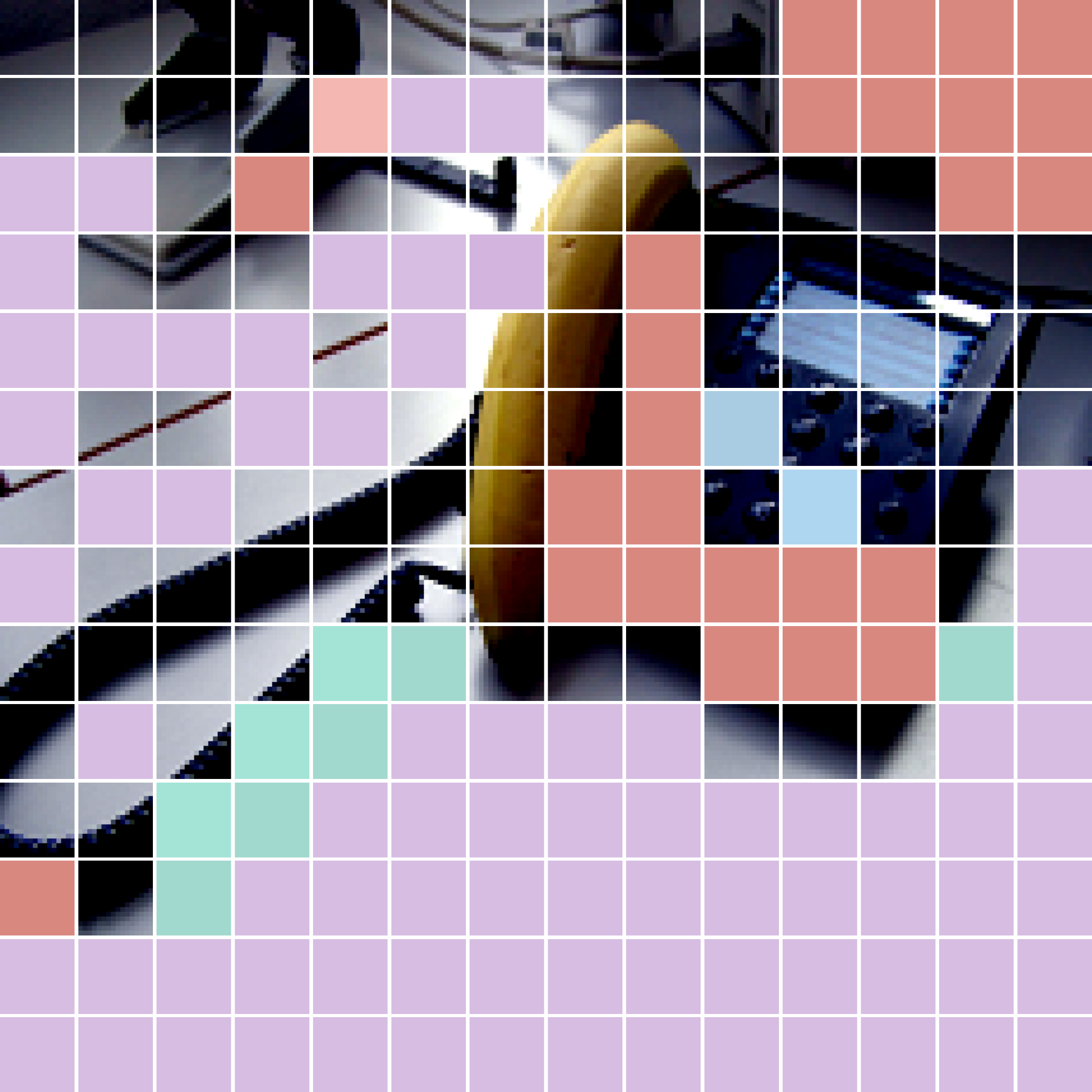}}
	\end{minipage}
    &
    \begin{minipage}[b]{0.3\columnwidth}
		\centering
		\raisebox{-.45\height}{\includegraphics[width=\linewidth]{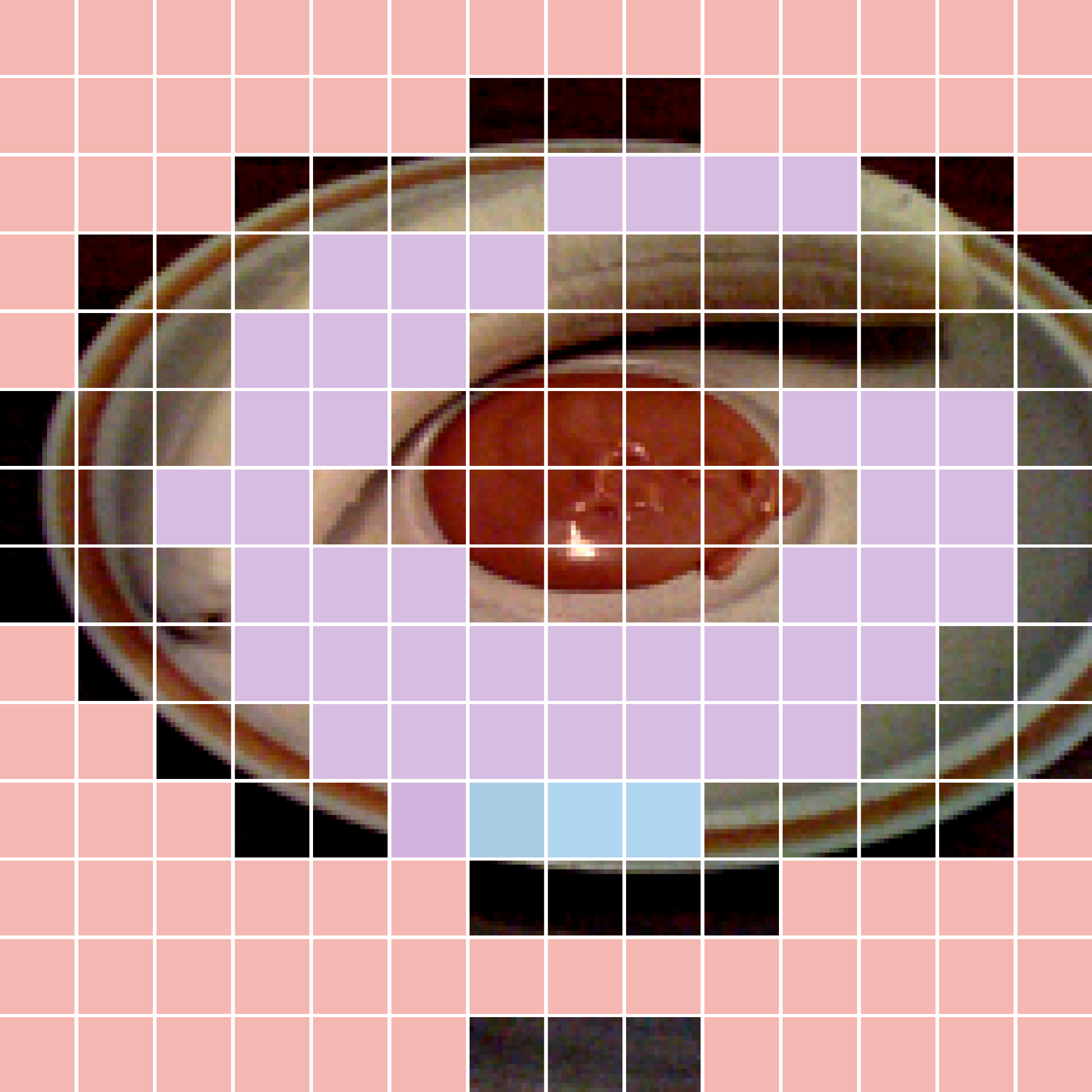}}
	\end{minipage}
    &
    \begin{minipage}[b]{0.3\columnwidth}
		\centering
		\raisebox{-.45\height}{\includegraphics[width=\linewidth]{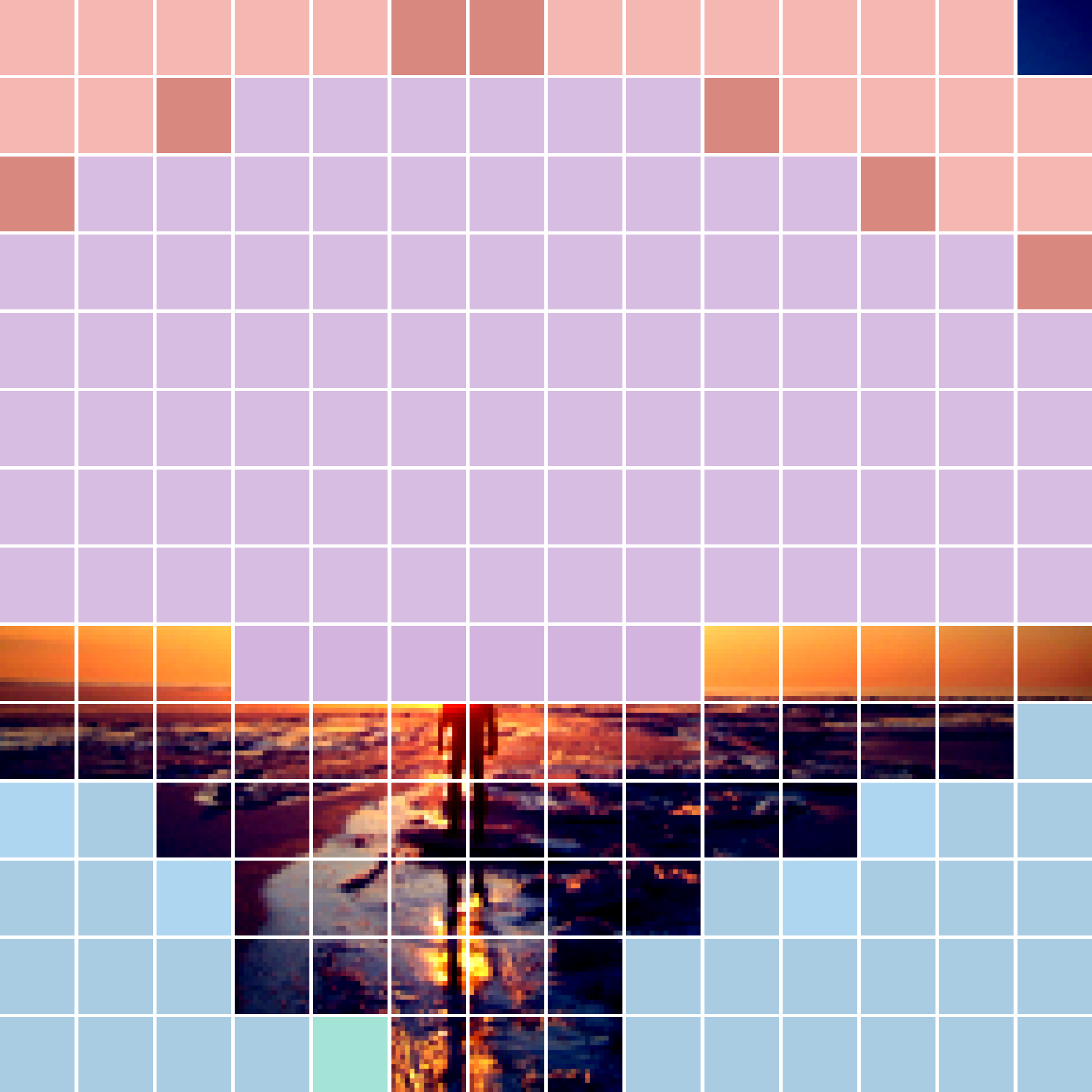}}
	\end{minipage}
    \\

    \rotatebox[origin=c]{90}{Caption} \hspace{-7pt}
    &\makecell{\small A large white \\ \small bowl of many \\ \small green apples.}
    &\makecell{\small A man flying \\ \small through the air \\ \small while riding skis.}
    &\makecell{\small A little girl \\ \small is holding an \\ \small  umbrella on a \\ \small wet day.}
    &\makecell{\small The telephone has \\ \small a banana where \\ \small the receiver \\ \small should  be.}
    &\makecell{\small A banana is \\ \small laying on a \\ \small small plate}
    &\makecell{\small A person standing \\ \small in shore of \\ \small beach with a \\ \small frisbee in the sky.}
    \\
  \end{tabular}
  }
  \endgroup
  \vspace{-2mm}
  \caption  
  {\textbf{Visualization of cluster masks.} Different colors represent distinct clusters formed by the similarity matrix calculated from the chosen anchor patches.} 
\label{fig:visual}
\vspace{-5mm}
\end{figure*}

\subsection{Main Results}
\paragraph{Visualization of Clusters.}
Figure \ref{fig:visual} offers a visual depiction of our cluster based masking technique as outlined in the methodology section. For this illustration, we randomly select a number of image-text pairs from the COCO validation set and apply our masking method to the pure RGB data of the images. The visualization is showing the masking result of the two-stages. In the first stage, a subset of patches (5\%) is randomly selected as anchor patches from the pool of all image patches, which are annotated with the red boxes. In the second stage, we visualize the masked clusters that are calculated based on the similarity matrix, where each cluster is represented by a distinct color.

\paragraph{Zero-shot Retrieval Results.}
In our investigation into the model's understanding of the relationship between visual and linguistic representations, we conduct zero-shot retrieval tests on several leading retrieval benchmarks. The results, detailed in Table \ref{tab:retrieval}, provide insights into the performance of our approach against others, particularly in the context of Image2Text and Text2Image's recall precision at top1(R1), top5(R5) and top10(R10) metrics.

In the evaluation on the MS-COCO \cite{lin2014microsoft}, Flickr8k, and Flickr30k \cite{flickr} datasets, our model outperforms both the baselines in most parts. Notably, in the Image-to-Text tasks, our model performs best in most datasets, with the exception of a slight performance decrease compared to $\text{FLIP}_\text{Attn}$ on the MS-COCO dataset. We attribute this success to our training strategy, which prioritizes primary clusters and minimizes the influence of noise. Furthermore, we observe that methods combining RGB information with token embeddings outperform those relying solely on RGB. We hypothesize that this is because the embedding layer, which contains slightly higher-level information. %

When comparing FLIP to CLIP, FLIP's performance is noticeably weaker, even with large batch sizes. We suspect that FLIP's sub-optimal results in our experimental settings may not fully exploit its strengths. This aligns with findings from other studies, such as Yang et al.'s research on ACLIP \cite{yang2022attentive}, which also noted FLIP's limitations. We observe that using attention scores for masking can improve performance compared to purely random masking. However, random masking still falls short of our cluster based masking or even the original CLIP method in some benchmarks. %

\begin{table*}[htbp]
\centering
\resizebox{\textwidth}{!}{
\begin{tabular}{lcccccccccccccccccc}
\multicolumn{1}{c}{}&\multicolumn{9}{c}{\textit{text $\rightarrow$ image}} & \multicolumn{9}{c}{\textit{image $\rightarrow$ text}} \\
\cmidrule(lr){2-10} \cmidrule(lr){11-19}
\multicolumn{1}{c}{}&\multicolumn{3}{c}{MS-COCO} & \multicolumn{3}{c}{Flickr8k} & \multicolumn{3}{c}{Flickr30k} & \multicolumn{3}{c}{MS-COCO} & \multicolumn{3}{c}{Flickr8k} & \multicolumn{3}{c}{Flickr30k}\\
\cmidrule(lr){2-4} \cmidrule(lr){5-7} \cmidrule(lr){8-10} \cmidrule(lr){11-13} \cmidrule(lr){14-16} \cmidrule(lr){17-19}
\multicolumn{1}{c}{}& R1 & R5 & R10 & R1 & R5 & R10 & R1 & R5 & R10 & R1 & R5 & R10 & R1 & R5 & R10 & R1 & R5 & R10\\
\midrule
    CLIP \cite{clip} & 34.60 & \textbf{61.98} & 72.72 & 55.70 & 81.60 & 89.90 & \textbf{58.50} & 83.80 & 89.10 & 23.49 & 47.80 & 59.66 & 40.54 & 68.90 & \textbf{80.20} & 43.18 & 70.44 & \textbf{80.40} \\
    FLIP \cite{flip} & 32.62 & 59.14 & 70.64 & 55.00 & 80.90 & 88.90 & 53.80 & 80.80 & 88.50 & 22.56 & 46.08 & 58.09 & 40.32 & 68.10 & 78.64 & 41.52 & 67.90 & 77.46 \\
    $\text{FLIP}_\text{Attn}$ \cite{yang2022attentive} & 33.66 & 60.18 & 71.02 & 53.70 & 80.09 & 87.99 & 55.29 & 81.40 & 87.60 & \textbf{23.89} & \textbf{48.33} & \textbf{60.04 }&  40.58 & 68.88 & 78.86 & 43.06 & 70.10 & 78.82  \\
     \hdashline
     $\text{Ours-Kmeans}$ & 33.68 & 61.14 & 71.97 & 55.10 & 83.30 & 90.90  & 55.40 & 82.00 & 88.90 & 22.60 &  46.93 &  58.84 &  40.10 & 69.86 & 79.91 & 41.32 & 68.50 & 77.58  \\
    $\text{Ours-RGB}_{0.5} $ & 32.82 & 60.20 & 71.40 & 52.10 & 81.20 & 89.50 & 54.90 & 81.20 & 88.00 & 22.97 & 47.29 & 59.21 & 40.84 & 68.72 & 79.14 & 41.80 & 70.20 & 79.42\\
    $\text{Ours-RGB}_{0.3} $ & \textbf{35.87} & 61.50 & 72.34 & 54.70 & 81.40 & \textbf{90.70} & 57.60 & 83.90 & \textbf{90.30} & 23.65 & 47.54 & 59.25 & 40.90 & 68.22 & 79.00 & 42.92 & 69.96 & 79.12 \\
    Ours-Embedding & 34.26 & 61.96 & \textbf{73.30} & \textbf{57.00} & \textbf{82.70} & 90.10 & 55.80 &\textbf{ 84.20} & 89.60 & 23.77 & 48.18 & 59.76 & \textbf{42.00} & \textbf{69.40} & 79.64 & \textbf{43.30} & \textbf{70.92} & 80.16 \\
\bottomrule
\end{tabular}
}
\vspace{-2mm}

\caption{\textbf{Zero-shot retrieval results.} We evaluate on MS-COCO \cite{lin2014microsoft}, Flickr8k and Flickr30k datasets \cite{flickr}, where the Recall@1 (R1), Recall@5 (R5), and Recall@10 (R10) are reported.}
\label{tab:retrieval}
\vspace{-2mm}
\end{table*}

\begin{table*}[htpb]
\centering
\resizebox{\textwidth}{!}{
\begin{tabular}{l|r|c|ccccccccccc|c}
\toprule
  \rotatebox[origin=l]{0}{Method} & \rotatebox[origin=l]{0}{N.T.\quad} &  \rotatebox[origin=l]{0}{IN-1K \cite{imagenet}} &    \rotatebox[origin=l]{0}{IN-A \cite{ImageNetA}} &    \rotatebox[origin=l]{0}{IN-O \cite{ImageNetO}}&    \rotatebox[origin=l]{0}{IN-R \cite{ImageNetR}} &  \rotatebox[origin=l]{0}{IN-S \cite{ImageNetS}} &  \rotatebox[origin=l]{0}{INv2 \cite{imagenetv2}} &  \rotatebox[origin=l]{0}{MNIST \cite{deng2012mnist}} &     \rotatebox[origin=l]{0}{Cal101 \cite{caltech101}}   &   \rotatebox[origin=l]{0}{CIFAR-10 \cite{krizhevsky2009learning}} &   \rotatebox[origin=l]{0}{CIFAR-100 \cite{krizhevsky2009learning}} & \rotatebox[origin=l]{0}{Flowers \cite{flower}} & \rotatebox[origin=l]{0}{Pets \cite{pets}}  &   \rotatebox[origin=l]{0}{Average} \\
\midrule
 CLIP \cite{clip}    &1.00$\times$ &  36.1    &       8.0 &           38.4 &         47.6 &      24.9            &   30.7      &   \textbf{11.7} &    73.5          &    57.7   &     25.0    & 26.0  & 53.9 & 36.1 \\
 FLIP \cite{flip}    &\textbf{0.53$\times$} &   34.4    &       7.1 &           39.5 &         41.4 &      20.1            &   29.5      &   10.4 &    70.4          &    52.8   &     24.5    & 25.3 & 46.0 & 33.5 \\

 $\text{FLIP}_\text{Attn}$ \cite{yang2022attentive} & 1.06$\times$ & 35.2 & 8.1 & 39.4 & 45.1 & 23.7 & 30.1 & 9.4 & 73.5 & 61.6 & 27.1  & 25.7 & 51.2 & 35.8\\
 \hdashline
$\text{Ours-Kmeans}$    &0.70$\times$ &   35.5    &       7.2 &           38.3 &         43.0 &      22.2            &   30.1      &   9.7 &    69.9          &    61.4   &     25.3   & 25.6 & 52.1 & 35.0 \\
 $\text{Ours-RGB}_{0.5} $ & 0.54$\times$ & 36.0 & 7.7 & 39.8 & 45.3 & 23.8 & 30.5& 11.2 & 72.2 & 63.6 & 27.2 & 26.1 & 55.0  & 36.5 \\
 $\text{Ours-RGB}_{0.3} $ & 0.64$\times$ & \textbf{36.6} & \textbf{8.8} & \textbf{39.9} & 45.9 & 24.9 & \textbf{31.8} & 9.4 & 72.3 & 63.1 & 26.3 & 25.4 & \textbf{57.3} & 36.8\\
 Ours-Embedding   & 0.64$\times$ & 36.3    & 8.1 & 39.6 & \textbf{47.9} & \textbf{25.4}            &   30.7     &   11.0 &   \textbf{73.7}          &   \textbf{ 70.7 }  &   \textbf{  32.0 }   & \textbf{28.4} &  55.4 & \textbf{38.2} \\
\bottomrule
\end{tabular}
}
\vspace{-2mm}
\caption{ \textbf{Zero-shot classification result.} We evaluate popular datasets using clip-benchmark \cite{clip_benchmark}. The training time is normalized according to the CLIP's training time. Here N.T. represents normalized time against and IN represents ImageNet.}
\label{zero_cls_result}
\vspace{-3mm}
\end{table*}

\begin{table}[htpb]
\centering
\resizebox{0.85\linewidth}{!}{
\begin{tabular}{lcccc}
   \toprule
   
   Method                       & CIFAR-10           & CIFAR-100              & IN-1K \\
   \midrule
   CLIP                         & 88.0              & 67.4                  & 62.3\\ 
   FLIP                         & 85.9              & 65.5                  & 61.3               \\ 
   $\text{FLIP}_\text{Attn}$ & 86.4 & 66.1 &  62.0\\
   \hdashline
   $\text{Ours-Kmeans}$ & 88.0 & 69.1 &  62.2\\
   $\text{Ours-RGB}_{0.5} $ & 86.7 & 66.2  & 62.5\\
   $\text{Ours-RGB}_{0.3} $                        & 88.6              & 68.7    & \textbf{63.1}               \\ 
   Ours-Embedding                       & \textbf{89.0}     & \textbf{69.7}         & 62.7 \\
   \bottomrule
\end{tabular}
}
\vspace{-2mm}
\caption{\textbf{Linear probing result.} All methods are trained for 10 epochs at learning rate of 1e-3. For CIFAR-10 and CIFAR-100, we use a batch size of 64 and for ImageNet-1k the batch size is 1024. }
\label{linear_probe_result}
\vspace{-5mm}
\end{table}

\paragraph{Results on Zero-shot Classification and Linear Probing.}
We evaluate our model on several widely recognized classification benchmarks. The zero-shot classification results are presented in Table \ref{zero_cls_result}, while the linear probing results can be found in Table \ref{linear_probe_result}. For better evaluating the time spent on training, we normalize all method's training time by the CLIP's training time, which is considered as $1\times$.

When comparing our model's performance to CLIP (i.e., no masking), our model demonstrates superior results on the majority of test cases, showcasing an average improvement of +\textbf{2.1\%}, with about +\textbf{36\%} speeding up. In comparison to the FLIP strategy, which has a similar training duration, our model has an improvement of +\textbf{5.5\%}. In comparison to the $\text{FLIP}_\text{Attn}$, our model does not need the attention map for guidance, which gives a much fast training speed, while having a performance of +\textbf{2.6\%} on average.

Out of 12 datasets on the zero-shot classification benchmark, our RGB and embedding model achieves the top performance on 11 of them.  In particular, it obtains strong performance on the ImageNet variants: ImageNet-A \cite{ImageNetA}, ImageNet-O \cite{ImageNetO}, ImageNet-R \cite{ImageNetO}, and ImageNet-S \cite{ImageNetS}, which often contain challenging and diverse images. The RGB version of our method also significantly outperforms FLIP and surpasses the CLIP model, especially on ImageNet and its variants, which demonstrates the effectiveness of our method with even the natural guidance.

The linear probing results further suggest the effectiveness of our method. 
Our models achieve +\textbf{1.8\%} accuracy on ImageNet, +\textbf{3.1\%} on CIFAR-10, and +\textbf{4.2\%} on CIFAR-100.

\begin{table*}[htpb]
\centering
\resizebox{0.8\linewidth}{!}{
\begin{tabular}{lcccccccccc}
\toprule
\multicolumn{1}{c}{}& \multicolumn{3}{c}{REPLACE} & \multicolumn{2}{c}{SWAP} & \multicolumn{2}{c}{ADD}  & \multicolumn{3}{c}{Average} \\
\cmidrule(lr){2-4} \cmidrule(lr){5-6} \cmidrule(lr){7-8}  \cmidrule(lr){9-11}
\multicolumn{1}{c}{}& Object & Attribute & Relation & Object & Attribute & Object & Attribute & Object & Attribute & Relation \\
\midrule
CLIP & 85.77 & \textbf{79.18} & 64.51 & 61.78 & 58.71 & 74.24 & 68.35 & 73.71 & 68.75 & 64.51 \\
FLIP & 84.07 & 75.88 & \textbf{66.00} & 60.16 & 61.56 & 71.67 & 63.15 & 71.97 & 66.86 & \textbf{66.00}\\
$\text{FLIP}_\text{Attn-0.5}$ &  86.62 & 75.50 & 63.22 &  52.44 & 63.06 & 71.65 & 66.76 & 71.57 & 68.39 & 63.22\\
$\text{FLIP}_\text{Attn-0.3}$ &  86.07 & 75.00 & 62.23 &60.57 & 59.16 & 74.68 & 68.64 & 72.82 & 63.47 & 62.23 \\
\hdashline
$\text{Ours-Kmeans}$ &  84.50 & 76.90 & 63.09 &62.20 & 61.86 & 71.77 & 65.46 & 73.66 & 67.60 & 63.09 \\
$\text{Ours-RGB}_{0.5}$ & \textbf{86.86} & 73.47 & 60.24 & 59.35 & 63.36 & 73.33 & 66.76 & 73.18 & 68.42 & 60.24 \\
$\text{Ours-RGB}_{0.3}$ & 86.13 & 75.13 & 64.65 & \textbf{66.67} & \textbf{63.36} & \textbf{74.92} & \textbf{71.24} & \textbf{75.91} & \textbf{69.91} & 64.65 \\
\bottomrule
\end{tabular}
}
\vspace{-2mm}
\caption{\textbf{Language composition test result.} This table presents the performance of models on the \sugarcrepe evaluation, which involves replacing, swapping, or adding atomic concepts such as objects, attributes, and relations in a sentence to create mismatched captions.}
\label{tab:composition}
\vspace{-5mm}
\end{table*}

\paragraph{Language Composition.}
A potential drawback of our method might be understanding compositions of concepts in language. As we mask out clusters,  there is a risk that the model may increasingly adopt \textit{bag-of-words} tendencies \cite{yuksekgonul2022and}, which could impede its ability to learn the relationships between objects. For example, if an image is captioned with "dog on grass", the grass may be masked for a large portion in our model as they are highly similar to each other. This will make learning the relation ``on'' difficult. Therefore, we apply \sugarcrepe \cite{hsieh2023sugarcrepe} benchmarks to test the model's ability to understand language compositions. \sugarcrepe benchmarks assess this by generating negative captions through manipulations like adding, swapping, or replacing concepts in sentences, followed by text retrieval tests to evaluate the model's accuracy in selecting the correct answer. From the our test results, which shown in Table \ref{tab:composition}, our model yields comparable results in Relation tests and demonstrates a significant enhancement in Object and Attribution tests, with an average improvement of \textbf{+3.9\%} and \textbf{+3.0\%} respectively, compared to FLIP. This improvement may stem from the masking of entire objects, which simplifies the challenge of contrastive learning by reducing ambiguity. This clarity facilitates the model's learning of relationships, a crucial factor for composition understanding.

\paragraph{Qualitative Comparison of Masking Strategies}
Our method outperforms the random masking strategy by preserving more semantic content in the unmasked image patches, a comparison showcased in Figure \ref{fig:teaser}. The advantage of our technique is further explored by the captioning experiment detailed in Figure \ref{fig:caption_visual}, wherein two sets of images, each masked differently, are fed into a captioner, GPT-4 \cite{openai2023gpt4, chatgpt}. The captioner is prompted to generate MS-COCO-style captions for the unobscured sections. When comparing these captions to the standard references, it becomes clear that our cluster based masking not only retains key elements but also the interrelations among them. For instance, our approach accurately enables the captioning system to identify an airplane in the first example and to describe the baseball player's action in the second, while the random masking strategy failed to achieve this clarity. These results indicate that our masking method provides a more detailed comprehension of the image.

\begin{figure}[htbp]
  \centering
   \resizebox{0.8\linewidth}{!}{
  \setlength{\tabcolsep}{1.2pt}{
  
  \begin{tabular}{ccc}
        Reference & Random Mask & Cluster Mask \\ 
        \begin{minipage}[b]{0.3\columnwidth}
		\centering
		{\raisebox{-.45\height}{\includegraphics[width=\linewidth]{fig/captioner/original_1.pdf}}}
	      \end{minipage}
        &
        \begin{minipage}[b]{0.3\columnwidth}
		\centering
		{\raisebox{-.45\height}{\includegraphics[width=\linewidth]{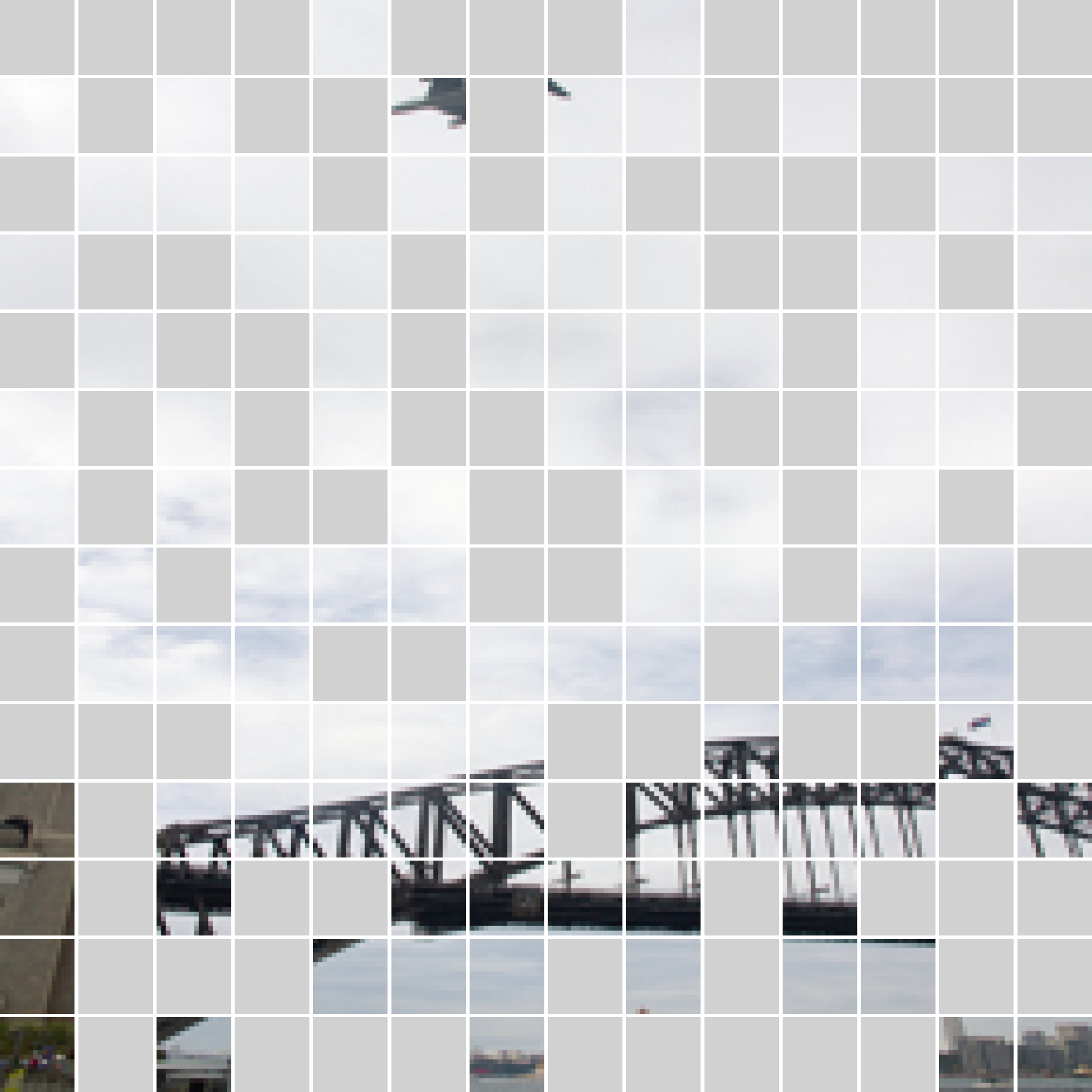}}}
	      \end{minipage}
        &
        \begin{minipage}[b]{0.3\columnwidth}
		\centering
		{\raisebox{-.45\height}{\includegraphics[width=\linewidth]{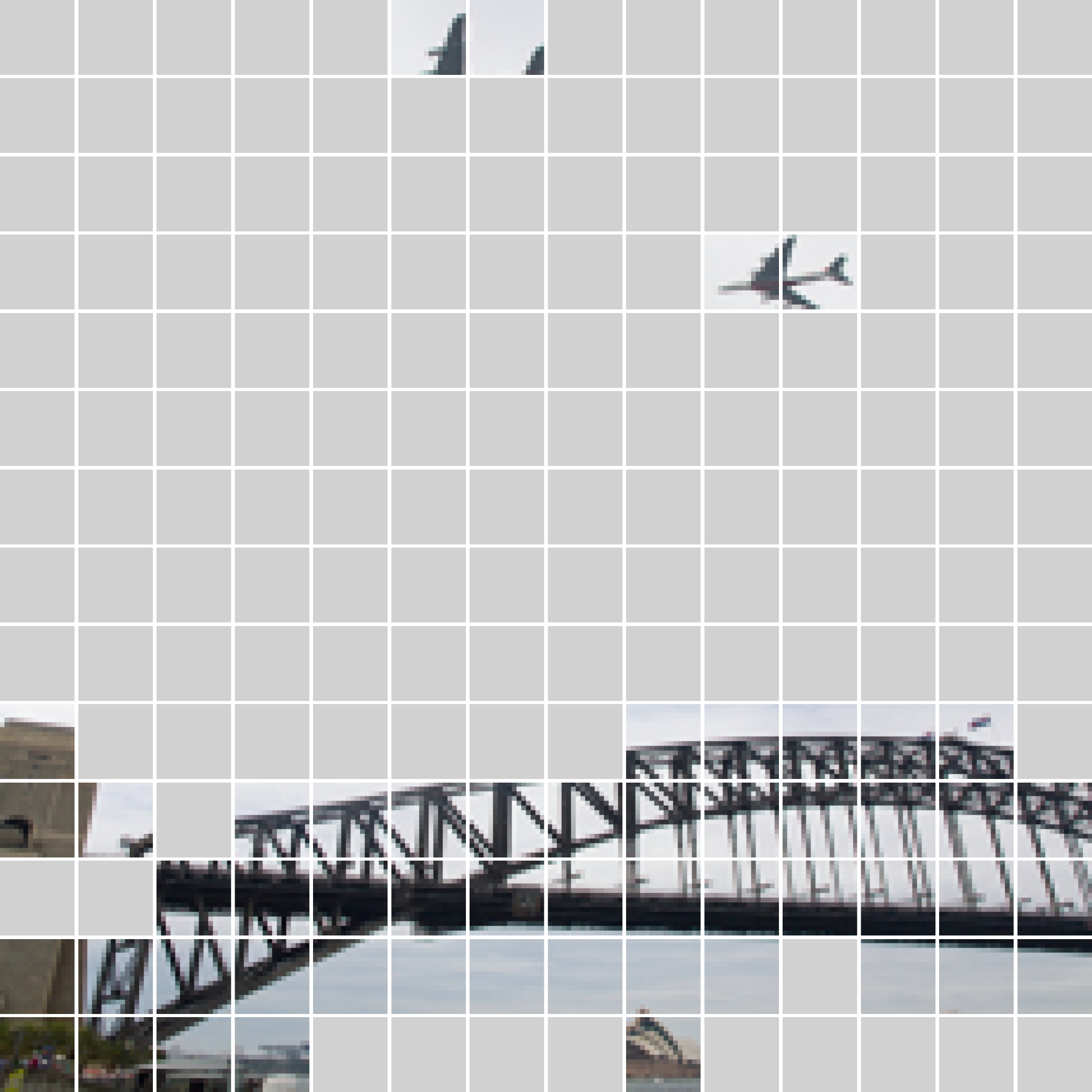}}}
	    \end{minipage}
     \\

     \vspace{-0.07\columnwidth}
     \makecell{\small \vspace{-3pt}  Two planes flying \\ \vspace{-3pt}  \small in the sky \\ \vspace{17pt} \small over a bridge. } 
    & \makecell{\small \vspace{-3pt} A clear sky \\ \vspace{-3pt}  \small above an \\ \vspace{17pt} \small arch bridge.} 
    & \makecell{\small \vspace{-3pt} Jets flying \\ \vspace{-3pt}  \small over an \\ \vspace{17pt} \small arch bridge. } \\
       \begin{minipage}[b]{0.3\columnwidth}
		\centering
		{\raisebox{-.45\height}{\includegraphics[width=\linewidth]{fig/captioner/original_2.pdf}}}
	      \end{minipage}
        &
        \begin{minipage}[b]{0.3\columnwidth}
		\centering
		{\raisebox{-.45\height}{\includegraphics[width=\linewidth]{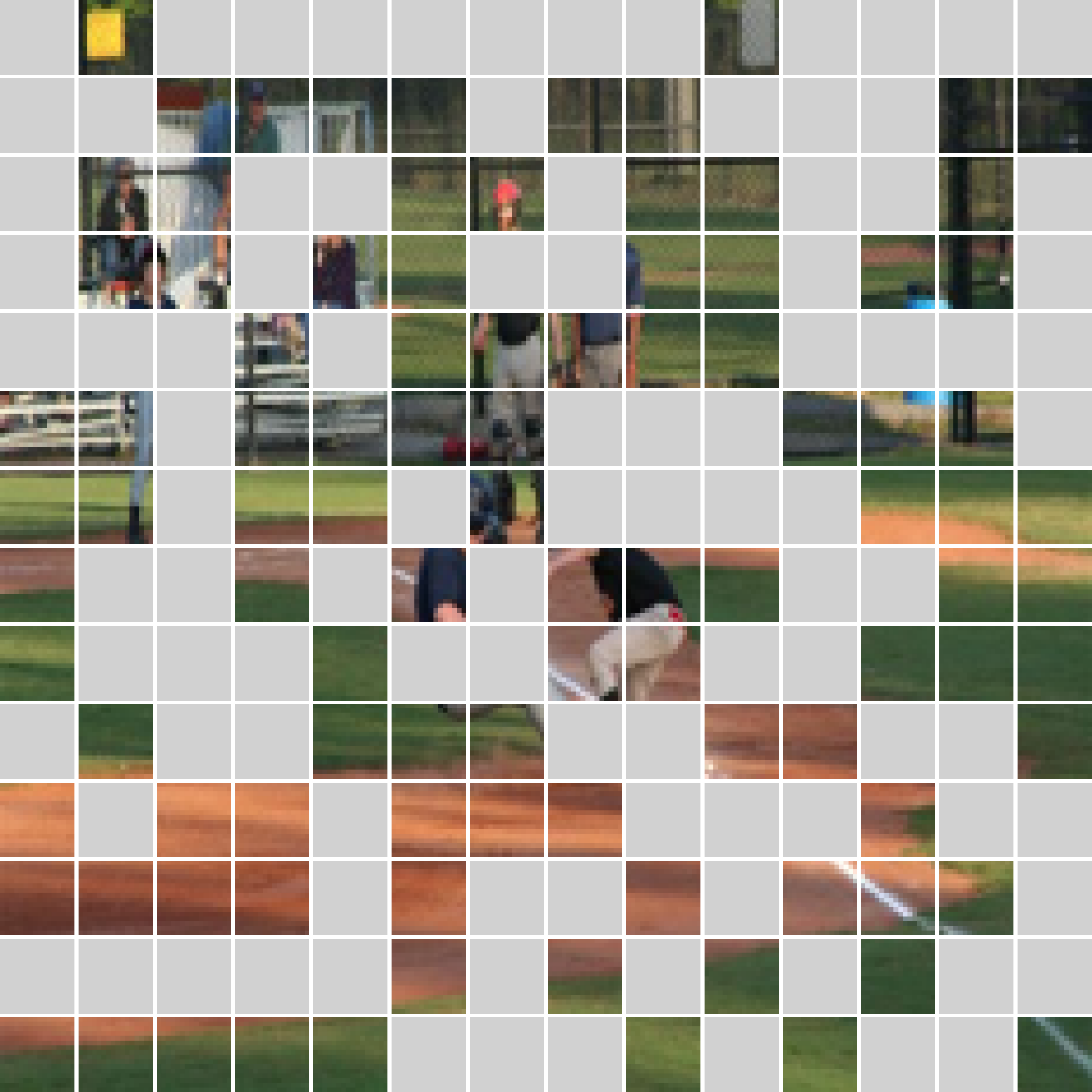}}}
	      \end{minipage}
        &
        \begin{minipage}[b]{0.3\columnwidth}
		\centering
		{\raisebox{-.45\height}{\includegraphics[width=\linewidth]{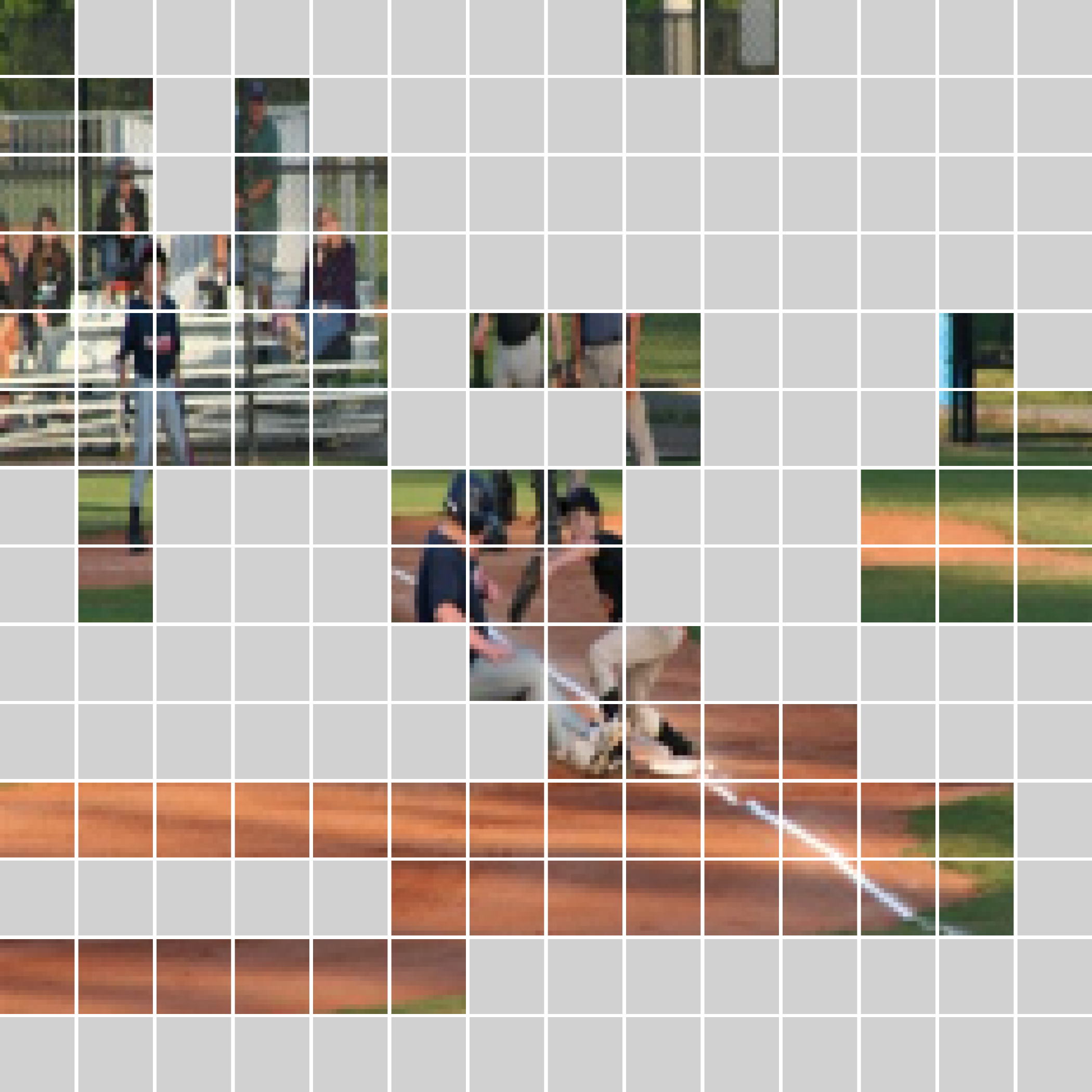}}}
	    \end{minipage}
     \\
    \vspace{-0.1\columnwidth}
     \makecell{\small \vspace{-3pt}  A boy catches a  \\\vspace{-3pt} \small ball as a player\\ \vspace{28pt}\small slides to the base.} 
    & \makecell{\small \vspace{-3pt} Players on a \\ \vspace{-3pt}  \small baseball field \\ \vspace{28pt} \small during a game.} 
    & \makecell{\small  \vspace{-3pt} A baseball player \\  \vspace{-3pt}\small sliding into home \\ \vspace{28pt}   \small {plate} during a game.} \\
  \end{tabular}
  }
  }
  \vspace{-2mm}
  \caption {\textbf{Generated caption from visible patches.} We process the masked images through GPT-4 \cite{openai2023gpt4, chatgpt} to create captions for the unmasked segments.} 
\label{fig:caption_visual}
\vspace{-5mm}
\end{figure}

\subsection{Ablation Study}

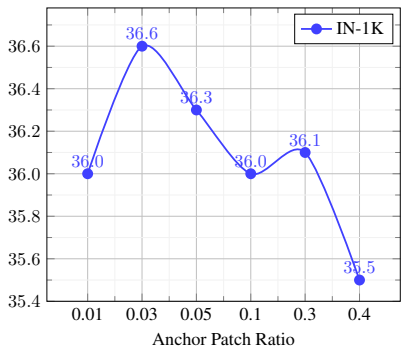
\begin{figure}[htbp]
\centering
\resizebox{0.65\linewidth}{!}{
\begin{tikzpicture}
\begin{axis}[
    xlabel={Anchor Patch Ratio},
    ymin=35.4,
    ymax=36.6,
    symbolic x coords={0.01, 0.03, 0.05, 0.1, 0.3, 0.4},
    xtick=data,
    nodes near coords,
    nodes near coords align={vertical},
    grid=both, %
    grid style={line width=.1pt, draw=gray!10}, %
    major grid style={line width=.2pt,draw=gray!50}, %
    minor tick num=1, %
    enlarge x limits=0.15, %
    enlarge y limits={upper,value=0.15}, %
    legend cell align={left},
    yticklabel style={/pgf/number format/.cd, fixed, fixed zerofill, precision=1},
    every node near coord/.append style={/pgf/number format/fixed zerofill, /pgf/number format/precision=1}, %
]

\addplot[
    color=blue!75, %
    mark=*, %
    mark size=2.5pt, %
    line width=1pt, %
    smooth, %
    ]
    coordinates {
    (0.01,36.0) (0.03,36.6) (0.05,36.3) (0.1,36.0) (0.3,36.1) (0.4,35.5)
    };
    \legend{IN-1K}

\end{axis}
\end{tikzpicture}
}

\vspace{-2mm}
\caption{\textbf{Effect of anchor patch ratio.} All the final masking ratio is tuned to be 50\%.}
\label{fig:ablation_anchor}
\vspace{-5mm}
\end{figure}

\paragraph{Ablation on Anchor Patch Ratio.}
In our study, we conducted an ablation on finding the optimal proportion of patches to serve as anchor patches. The results of this ablation are summarized in Figure \ref{fig:ablation_anchor}. We use zero-shot learning results on the ImageNet-1k dataset as the benchmark for assessing the quality of the representations learned by our model. Additionally, we calibrate the threshold for each experiment to ensure that the average final masking ratio was maintained at 50\%.

Our findings indicate that a smaller proportion of anchor patches tends to yield superior performance. We hypothesize that this improvement is due to the decreased randomness in the selection of anchor patches, which in turn enhances the clustering performance by providing a more stable set of reference points for the model to learn from. However, the masking ratio cannot be too small, as the similarity threshold will become too small, which may reduce the clustering quality. 

\paragraph{Ablation on Minimum Mask Ratio.}
We further demonstrate the capability of our method by setting the minimum mask ratio $\beta$ the same as the average masking ratio of FLIP, as shown in Table \ref{tab:retrieval}, \ref{zero_cls_result}, and Table \ref{tab:ablation_cutoff_ratio}. For our method, the cutoff ratio denotes the minimum mask ratio applied and the true visible patch ratio is shown in visible ratio. For the FLIP counterpart, it maintains a consistent mask ratio across all images.
The results indicate that our method not only matches the speed of FLIP but also surpasses FLIP in zero-shot ImageNet-1K classification accuracy with a +\textbf{1.6}\% improvement even by seeing fewer patches. The attention based masking method with less mask ratio achieves similar perfromance as ours but the speed is much slower.

These findings suggest that cluster based masking serves as an effective denoising technique for the dataset. A reason for this enhanced performance is that we could easily mask out typically irrelevant areas, such as uniformly colored backgrounds, which are less informative and oftentimes do not correspond to any word in the caption. This targeted approach enables the model to focus on more meaningful content within the images.

Additionally, our findings reveal an improved feature learning by the model when a smaller random masking is applied. By reducing the cutoff ratio from 50\% to 30\%, we observed a 1\% enhancement in classification accuracy.  Thus,  there is some trade-off between the model's performance and speed. Despite this, our model with larger masking cut off still remains significantly faster compared to attention-based masking or the original CLIP method. 

\begin{table}[htpb]
\centering
\resizebox{0.9\linewidth}{!}{
\begin{tabular}{lccccc}
   \toprule
   Method                       & $\beta$    & Visible Ratio     & Normalized Time              & IN-1K \\
   \midrule
   FLIP                         & 50\%       & 50\%   &       $0.84 \times$             &     34.4         \\ 
   FLIP                         & 30\%       &  70\%  &       $1.00 \times$            &        35.4       \\ 
   $\text{FLIP}_{\text{attn}}$           & 50\%       &  50\%  &       $1.73 \times$            &        35.2      \\ 
   $\text{FLIP}_{\text{attn}}$           & 30\%       &  70\%  &       $1.97 \times$            &        36.6 \\ 
   Ours-RGB                         & 50\%    &  43\%    &       $0.84 \times$             &     36.0          \\ 
   Ours-RGB                         & 30\%    &  50\%    &       $1.00 \times$             &     36.6           \\ 
   \bottomrule
\end{tabular}
}
\vspace{-2mm}
\caption{\textbf{Ablation on minimum mask ratio $\beta$}. Comparison of various methods against different minimum masking ratios. The zero-shot ImageNet-1k classification results are used as metric. The time is normalized to ours RGB model with $\beta$=30\%.}
\label{tab:ablation_cutoff_ratio}
\vspace{-5mm}
\end{table}

\paragraph{Ablation on Pixel Normalization.}
In our experiments, we incorporate pixel normalization (making each patch mean zero and unit standard deviation 1) into the process of computing the similarity matrix for images. This yields a performance improvement of +1.1\%, as shown in the results presented in Table \ref{tab:pixel_norm}. The underlying rationale for this enhancement is attributed to the standardization of image patches. By using pixel normalization, we focus on the relative intensity of pixels, thereby diminishing the impact of lighting variations among different images. %

This normalization process is particularly beneficial in scenarios where the dynamic range of pixel values varies significantly across different patches. By scaling the patches to a common range, pixel-norm mitigates the risk of disproportionate influence from patches with higher intensity values. Consequently, this leads to a more balanced and equitable comparison among patches, enhancing the model's ability to discern and quantify similarities more effectively.

\begin{table}[htpb]
\centering
\begin{subtable}{.45\linewidth}
\centering
\resizebox{0.75\linewidth}{!}{
\begin{tabular}{lc}
   \toprule
   Method & IN-1K \\
   \midrule
   w/o P.N. & 35.5 \\
   w/\text{ }\text{ } P.N. & \textbf{36.6} \\ 
   \bottomrule
\end{tabular}
}
 \vspace{11pt}
\caption{Ablation on pixel normalization (P.N.). }
\label{tab:pixel_norm}
\end{subtable}
\hfill
\begin{subtable}{.45\linewidth}
\centering
\resizebox{0.6\linewidth}{!}{
\begin{tabular}{lc}
   \toprule
   $k$ & IN-1K \\
   \midrule
   0.5 & 36.1 \\ 
   1 & \textbf{36.3} \\ 
   2 & 35.9 \\ 
   \bottomrule
\end{tabular}
}
\caption{Ablation on polynomial coefficient $k$.} 
\label{tab:combine}
\end{subtable}
\vspace{-2mm}
\caption{\textbf{Ablation Study.} Table \ref{tab:pixel_norm} presents an ablation study on the application of pixel normalization when calculating the similarity matrix for clustering. Table \ref{tab:combine} explores the effects of varying the polynomial coefficient \( k \), which adjusts the adaptive rate used when combining RGB and embedding features.}
\vspace{-5mm}
\end{table}

\begin{figure}[htbp]
  \centering
  \setlength{\tabcolsep}{2pt}{
  \resizebox{0.85\linewidth}{!}{
  \begin{tabular}{ccc}
        \begin{minipage}[b]{0.3\columnwidth}
		\centering
		{\includegraphics[width=\linewidth]{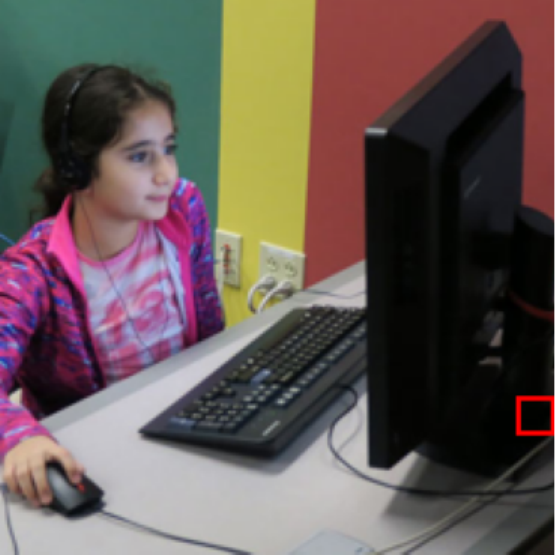}}
	      \end{minipage}
    & 
        \begin{minipage}[b]{0.3\columnwidth}
		\centering
		{\includegraphics[width=\linewidth]{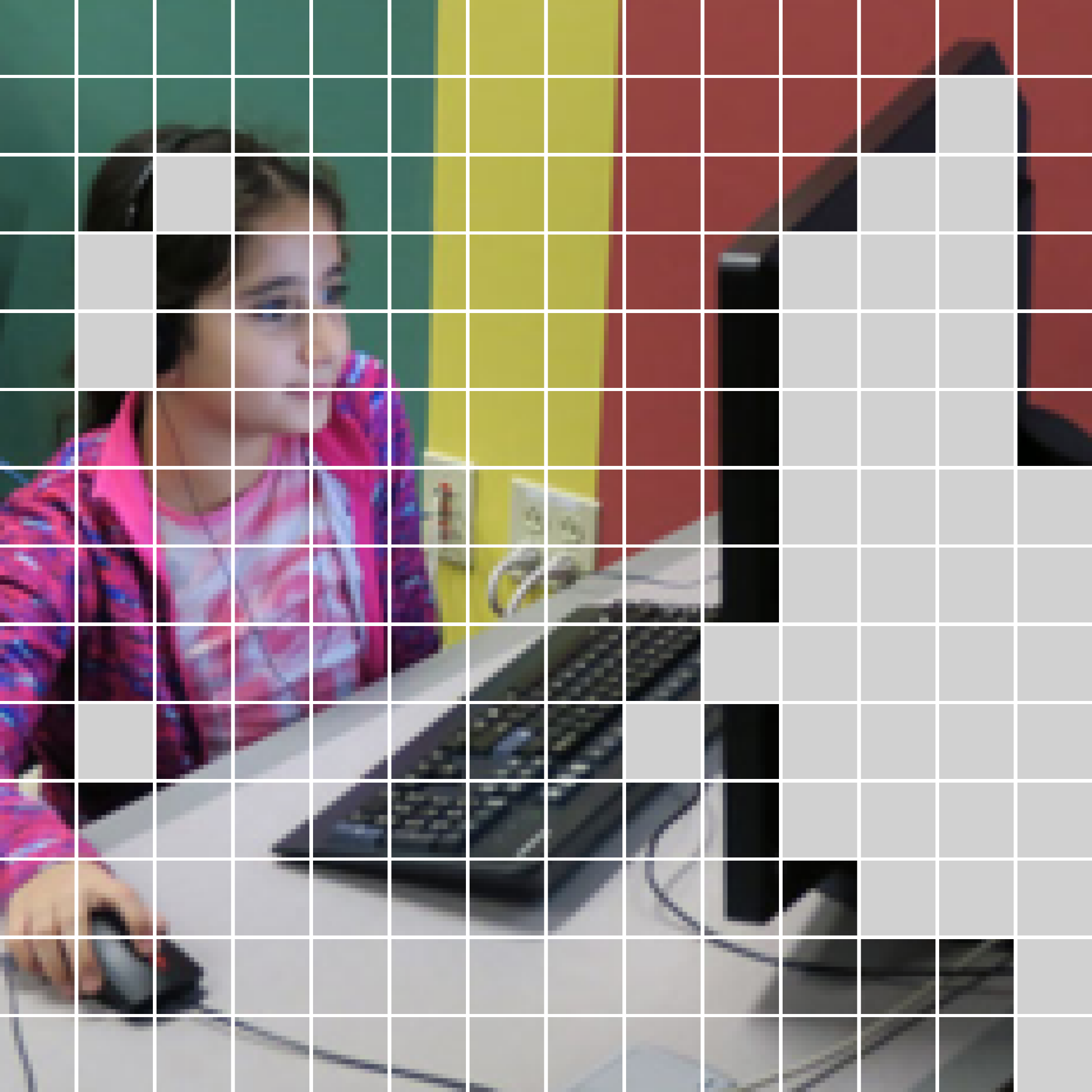}}
	      \end{minipage}
    &
        \begin{minipage}[b]{0.3\columnwidth}
		\centering
		{\includegraphics[width=\linewidth]{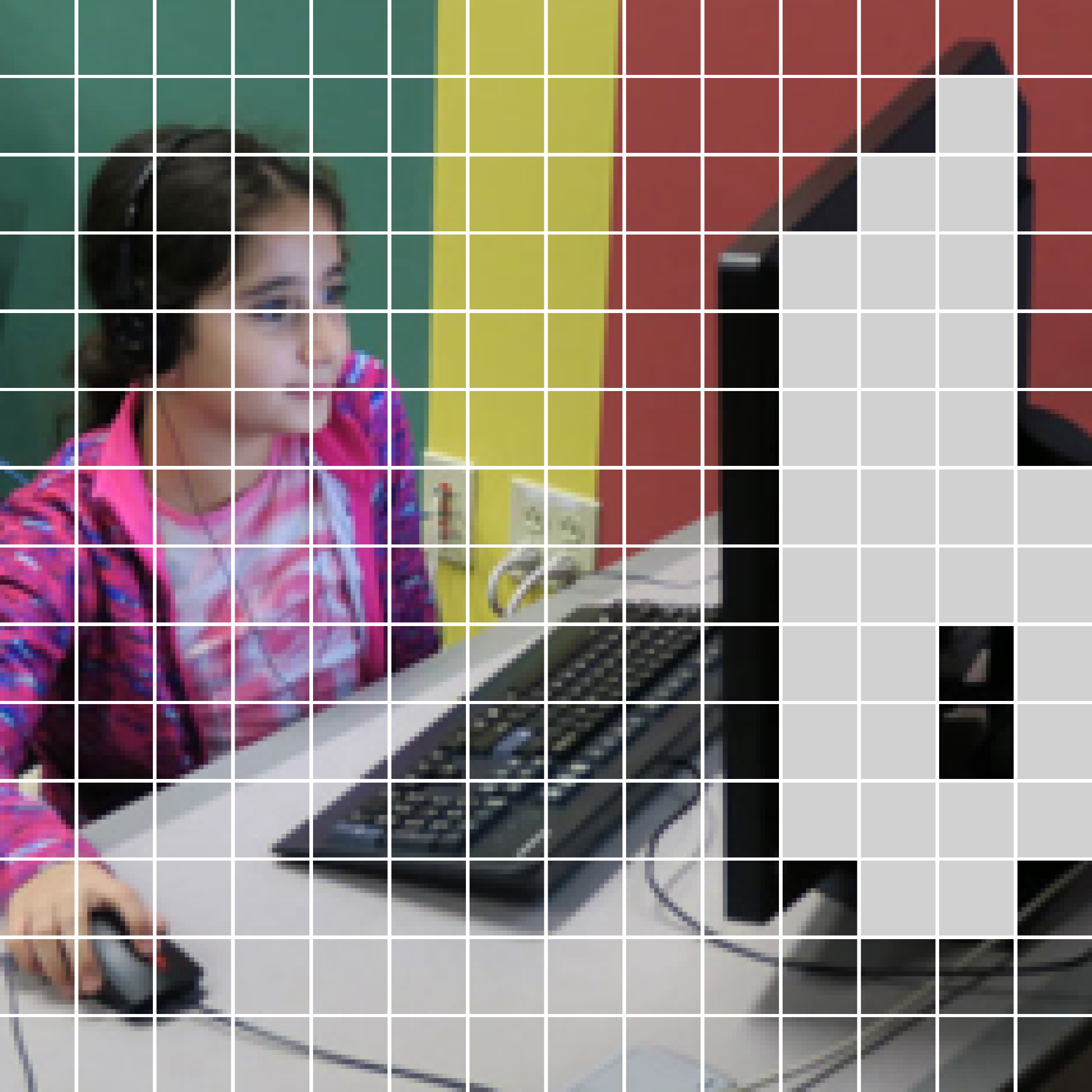}}
	    \end{minipage}
    \\
    \begin{minipage}[b]{0.3\columnwidth}
		\centering
		{\includegraphics[width=\linewidth]{fig/feature/anchor_chosen_2.pdf}}
	      \end{minipage}
    & 
        \begin{minipage}[b]{0.3\columnwidth}
		\centering
		{\includegraphics[width=\linewidth]{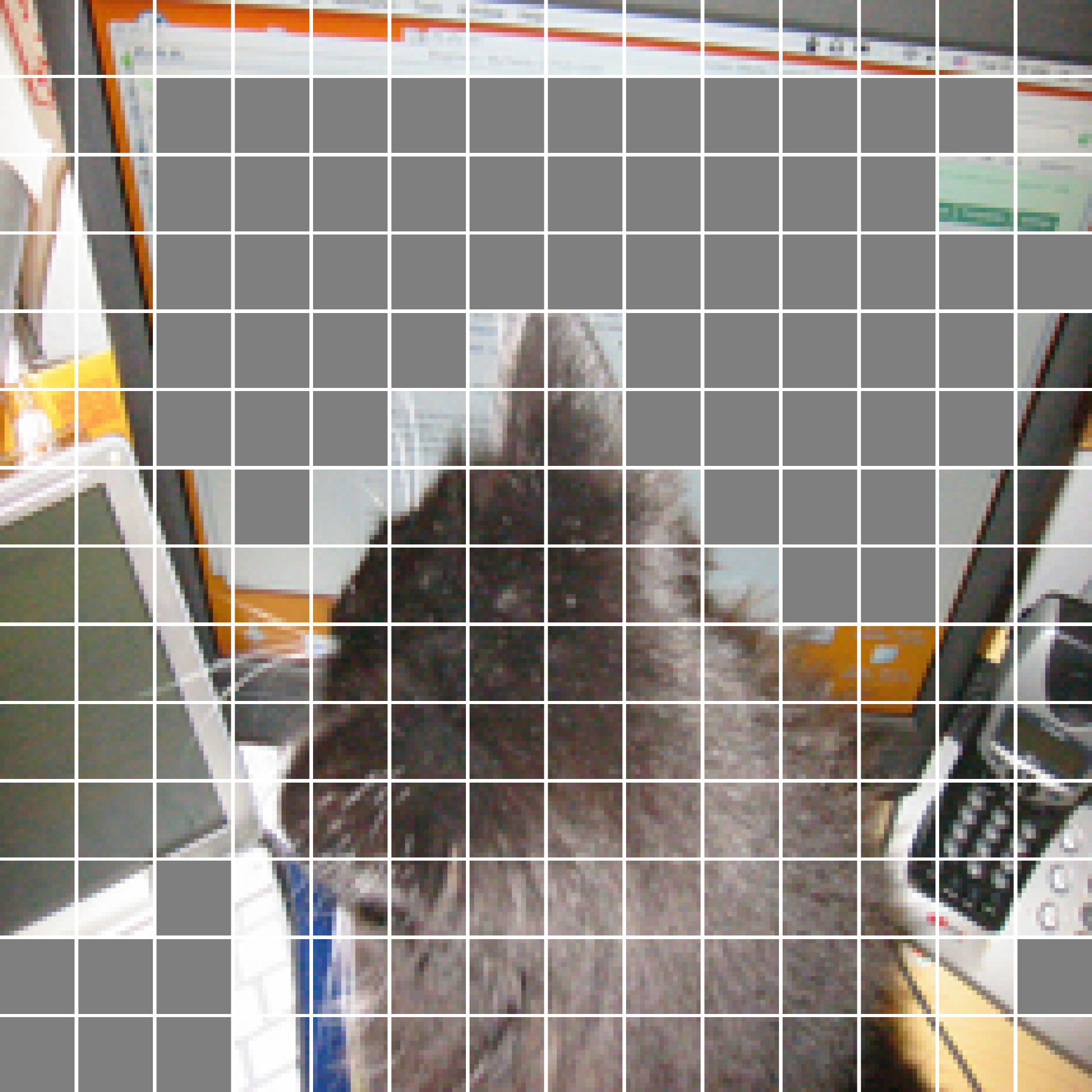}}
	      \end{minipage}
    &
        \begin{minipage}[b]{0.3\columnwidth}
		\centering
		{\includegraphics[width=\linewidth]{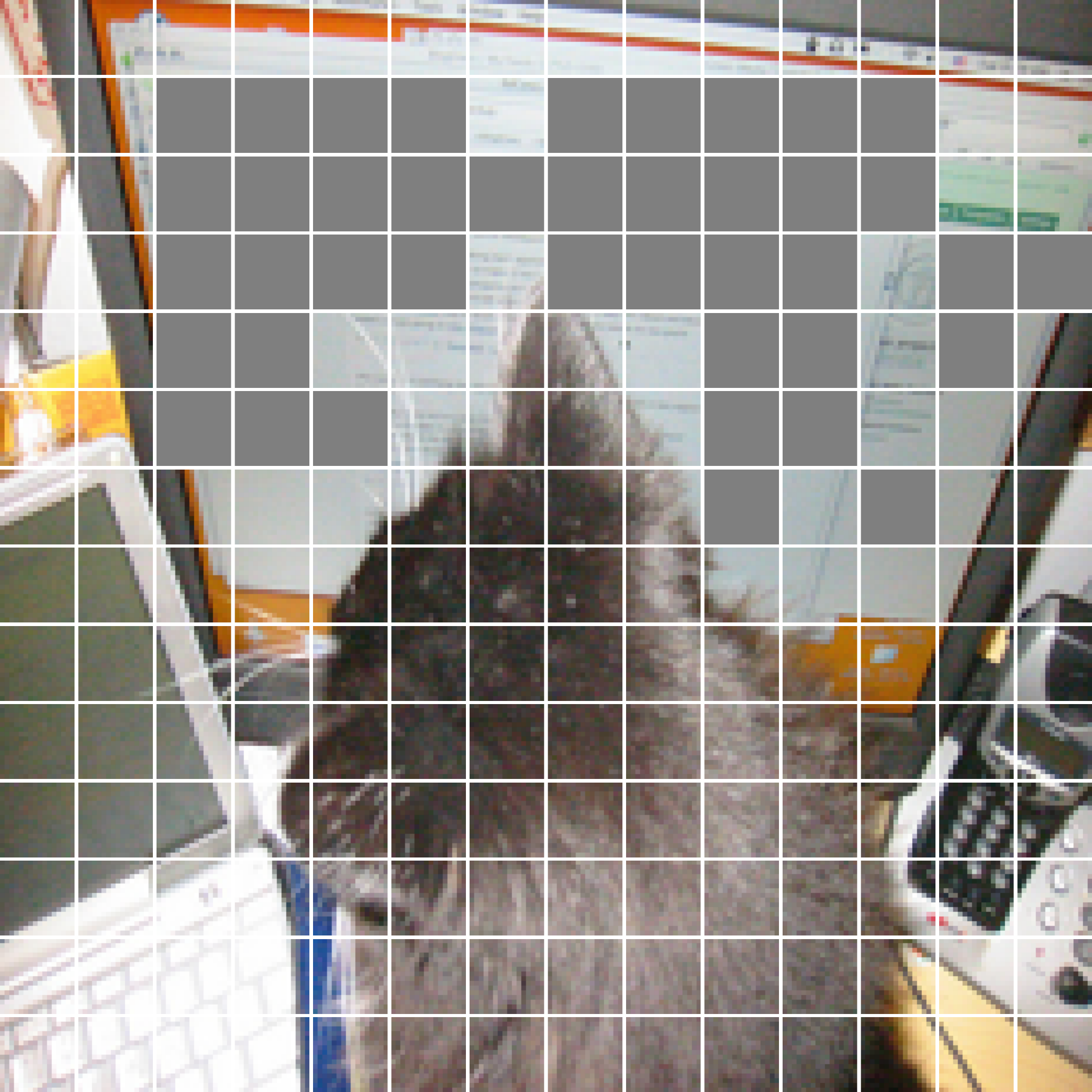}}
	    \end{minipage}
    \\
    Anchor Patch & RGB-based & Embedding-based\\ 
  \end{tabular}
  }
  }
  \vspace{-2mm}
    \caption {\textbf{Representation used in clustering.} We illustrate the distinctions between using pure RGB and ViT embeddings to compute the similarity matrix used in masking patches.}

\label{fig:ablation_feature}
\vspace{-5mm}
\end{figure}

\paragraph{Effect of Features used in Clustering.}
In Tables \ref{tab:retrieval} and \ref{zero_cls_result}, embedding-based methods surpass those dependent solely on RGB data, especially in image-to-text retrieval tasks. One reason for this may be the fact that the embedding model has access to the positional encoding, whereas the RGB-based model solely uses the appearance of each patch.  %
Figure \ref{fig:ablation_feature} qualitatively shows this advantage: while the RGB-only approach masks extra areas (such as hair or shadows in the first scenario; a laptop and phone in the second) due to color similarities, the embedding-based method masks more complete object parts. %

\paragraph{Ablation on Adaptive Rate.}
In our approach, we interpolate between using RGB features and the patch embedding layer feature using a coefficient, $\alpha$, which varies with each epoch. Denoting the current running epoch as $\mathrm{E}_{c}$ and the number of total training epochs as $\mathrm{E}_{t}$, this coefficient is defined as  $\alpha=\left(\frac{\mathrm{E}_{c}}{\mathrm{E}_{t}}\right)^k$. In addition to the linear method where $k=1$, we explore other polynomial coefficients $k$ for this combination, as summarized in Table \ref{tab:combine}. Our findings suggest that the linear combination is most effective, likely due to its smooth transition.

\subsection{Limitations}

Our methodology uses a uniform threshold for all images, a strategy that, while effective, may not be the most optimal. Future research could explore the implementation of individualized thresholds for each image, potentially leading to a more intelligent and adaptive masking process. 

All of our approaches use the popular backbone architecture ViT-B/16 \cite{dosovitskiy2020vit} and are trained solely on the CC12M dataset \cite{cc12m}. Expanding the scope of the experiments could offer additional insights.%

\section{Conclusion}
In our study, we introduce a novel cluster based masking strategy designed for vision-language pret-raining. Using either pure RGB values or shallow features from the patch embedding layer, our method effectively clusters image patches, maintaining essential visual semantics. We then randomly mask out these clusters, enabling efficient training. Our approach demonstrates success across various downstream evaluation tasks, including both pure image-based tasks such as image classification and multimodal tasks like image-text retrieval and language composition tests. We believe our work marks a considerable progression in this domain and anticipate that it will stimulate further research into optimizing masking strategies for similar applications.

\paragraph{Author Contributions}
All authors contributed to designing projects, launching experiments, and writing the paper. Zixuan Pan focused on algorithm optimization; Zihao Wei focused on code design; Andrew Owens supervised this project, offered feedback, and assisted in writing the paper.

\paragraph{Acknowledgements} This research is supported by a Sony Research Award. We are grateful to Jeongsoo Park, Chao Feng, Yiming Dou, Daniel Geng, Ziyang Chen, and Ayush Shrivastava for their valuable suggestions and discussion.

{
    \small
    \bibliographystyle{ieeenat_fullname}
    \bibliography{main}
}

\newpage
\appendix
\section{Qualitative Comparison of Masking Strategy}
Our method outperforms a random masking strategy by preserving more semantic content in the unmasked image patches, a comparison showcased in Figure \ref{fig:teaser}. The advantage of our technique is underscored by the captioning experiment detailed in Figure \ref{fig:caption_visual}, wherein two sets of images, each masked differently, were fed into a captioner, GPT-4 \cite{openai2023gpt4, chatgpt}. The model was tasked to generate MSCOCO-style captions for the unobscured sections. When comparing these captions to the standard references, it becomes clear that our cluster based masking not only retains key depicted elements but also the interrelations among them. For instance, our approach accurately enabled the captioning system to identify an airplane in the first example and to describe the baseball player's action in the second, while the random masking strategy failed to achieve this clarity. These results indicate that our masking method provides a more detailed comprehension of the image.

\begin{figure}[htbp]
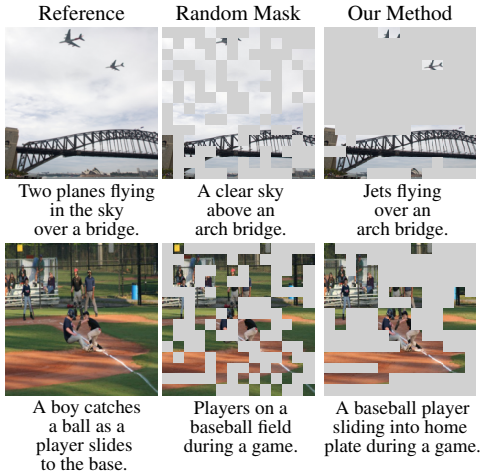

  \centering
   \resizebox{0.8\linewidth}{!}{
  \setlength{\tabcolsep}{1.2pt}{
  
  \begin{tabular}{ccc}
        Reference & Random Mask & Our Method \\ 
        \begin{minipage}[b]{0.3\columnwidth}
		\centering
		{\raisebox{-.45\height}{\includegraphics[width=\linewidth]{fig/captioner/original_1.pdf}}}
	      \end{minipage}
        &
        \begin{minipage}[b]{0.3\columnwidth}
		\centering
		{\raisebox{-.45\height}{\includegraphics[width=\linewidth]{fig/captioner/random_masking_1.pdf}}}
	      \end{minipage}
        &
        \begin{minipage}[b]{0.3\columnwidth}
		\centering
		{\raisebox{-.45\height}{\includegraphics[width=\linewidth]{fig/captioner/our_masking_1.pdf}}}
	    \end{minipage}
     \\

     \vspace{-0.07\columnwidth}
     \makecell{\small \vspace{-3pt}  Two planes flying \\ \vspace{-3pt}  \small in the sky \\ \vspace{17pt} \small over a bridge. } 
    & \makecell{\small \vspace{-3pt} A clear sky \\ \vspace{-3pt}  \small above an \\ \vspace{17pt} \small arch bridge.} 
    & \makecell{\small \vspace{-3pt} Jets flying \\ \vspace{-3pt}  \small over an \\ \vspace{17pt} \small arch bridge. } \\
       \begin{minipage}[b]{0.3\columnwidth}
		\centering
		{\raisebox{-.45\height}{\includegraphics[width=\linewidth]{fig/captioner/original_2.pdf}}}
	      \end{minipage}
        &
        \begin{minipage}[b]{0.3\columnwidth}
		\centering
		{\raisebox{-.45\height}{\includegraphics[width=\linewidth]{fig/captioner/random_masking_2.pdf}}}
	      \end{minipage}
        &
        \begin{minipage}[b]{0.3\columnwidth}
		\centering
		{\raisebox{-.45\height}{\includegraphics[width=\linewidth]{fig/captioner/our_masking_2.pdf}}}
	    \end{minipage}
     \\
    \vspace{-0.1\columnwidth}
     \makecell{\small \vspace{-3pt}  A boy catches \\ \vspace{-3pt}  \small a ball as a \\ \vspace{-3pt}\small player slides \\ \vspace{20pt}\small to the base.} 
    & \makecell{\small \vspace{-3pt} Players on a \\ \vspace{-3pt}  \small baseball field \\ \vspace{28pt} \small during a game.} 
    & \makecell{\small  \vspace{-3pt} A baseball player \\  \vspace{-3pt}\small sliding into home \\ \vspace{28pt}   \small {plate} during a game.} \\
  \end{tabular}
  }
  }
  \caption {\textbf{Generated caption from visible patches.} We process the masked images through GPT-4 \cite{openai2023gpt4, chatgpt} to create captions for the unmasked segments.} 
\label{fig:caption_visual}
\end{figure}

\section{Visualization of attention-based masking}
\begin{figure}[htbp]
  \centering
  \begingroup
  \setlength{\tabcolsep}{2pt}
  \resizebox{0.8\linewidth}{!}{
  \begin{tabular}{cccc}
    \small{Attn. Init.} & \small{Attn. E16} & \small {Attn. E32} & \small{Ours} \\
    \vspace{0.018\columnwidth}
    \begin{minipage}[b]{0.25\columnwidth}
		\centering
		\raisebox{-.45\height}{\includegraphics[width=\linewidth]{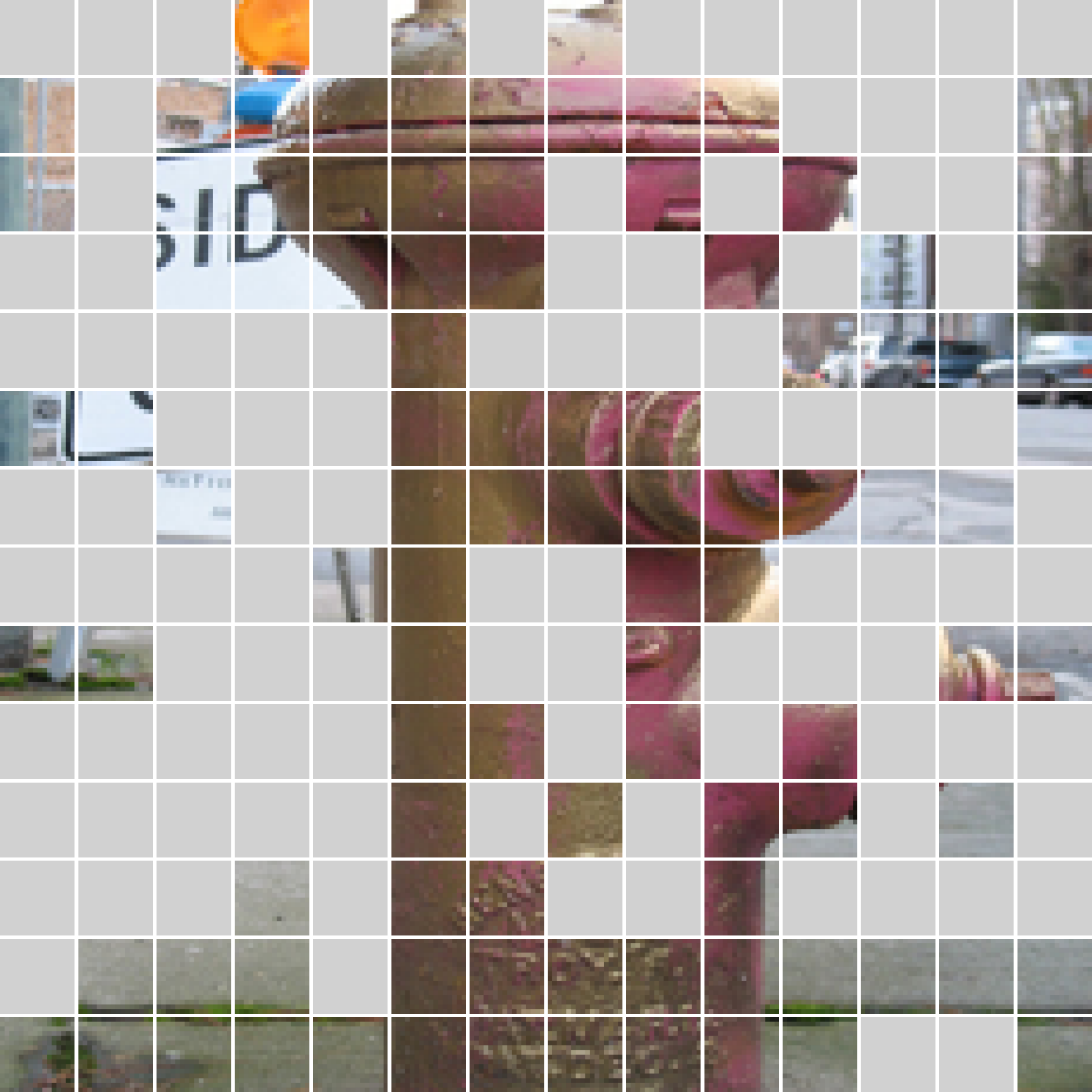}}
	\end{minipage}
    & 
    \begin{minipage}[b]{0.25\columnwidth}
		\centering
		\raisebox{-.45\height}{\includegraphics[width=\linewidth]{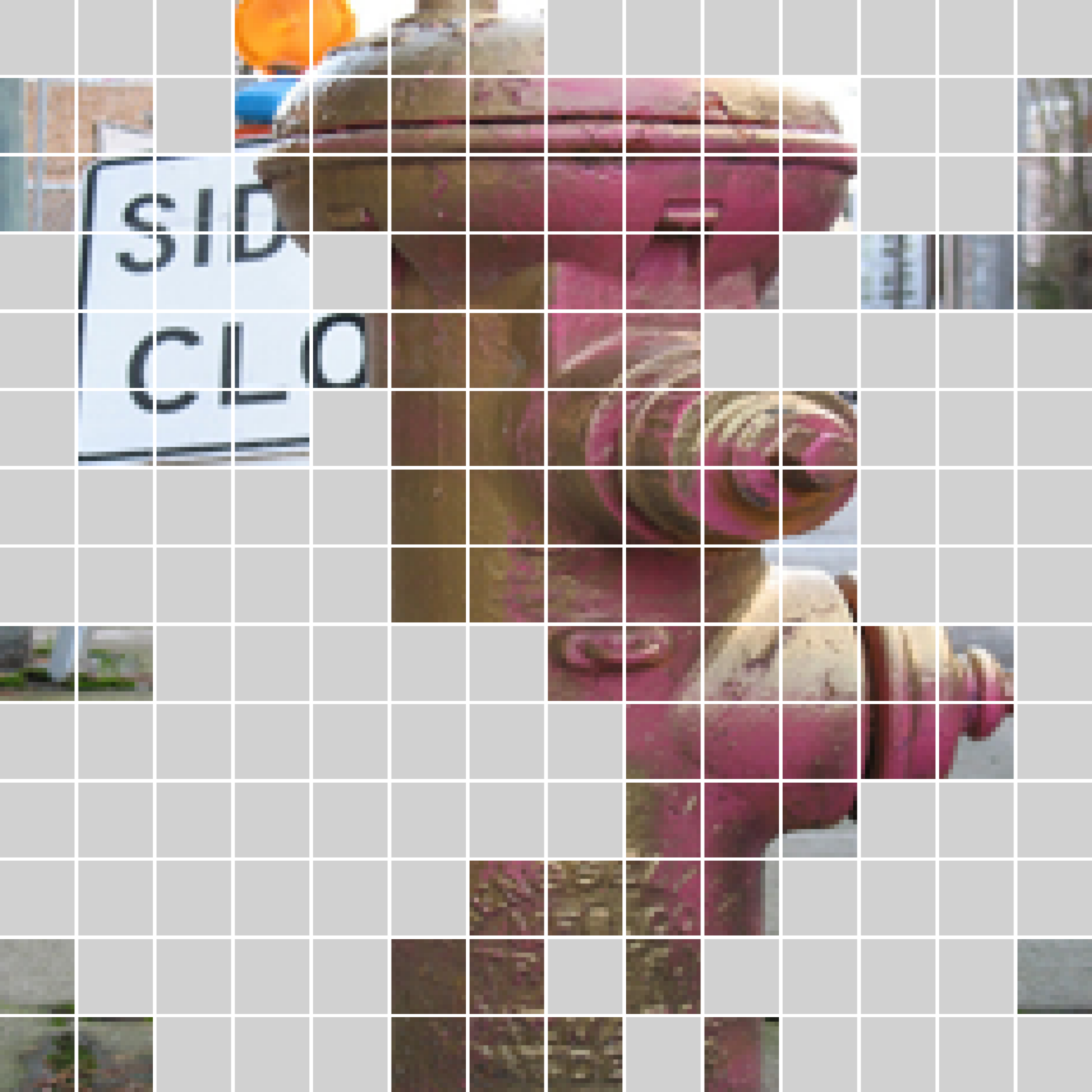}}
	\end{minipage}

   &    \begin{minipage}[b]{0.25\columnwidth}
		\centering
		\raisebox{-.45\height}{\includegraphics[width=\linewidth]{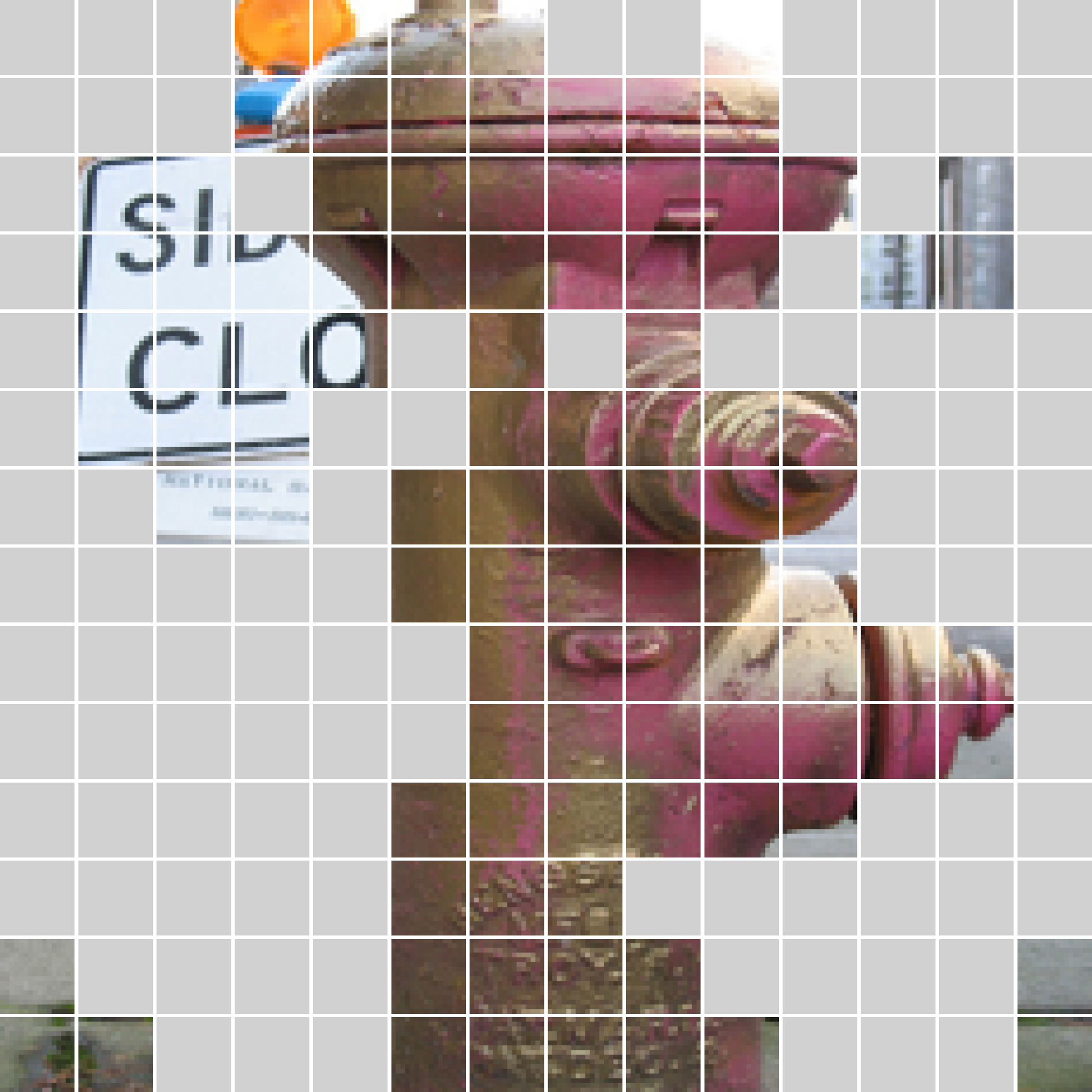}}
	\end{minipage}
 &    \begin{minipage}[b]{0.25\columnwidth}
		\centering
		\raisebox{-.45\height}{\includegraphics[width=\linewidth]{fig/teaser/teaser_cluster_masking_1.pdf}}
	\end{minipage}
    
    \\

    \vspace{0.018\columnwidth}

        \begin{minipage}[b]{0.25\columnwidth}
		\centering
		\raisebox{-.45\height}{\includegraphics[width=\linewidth]{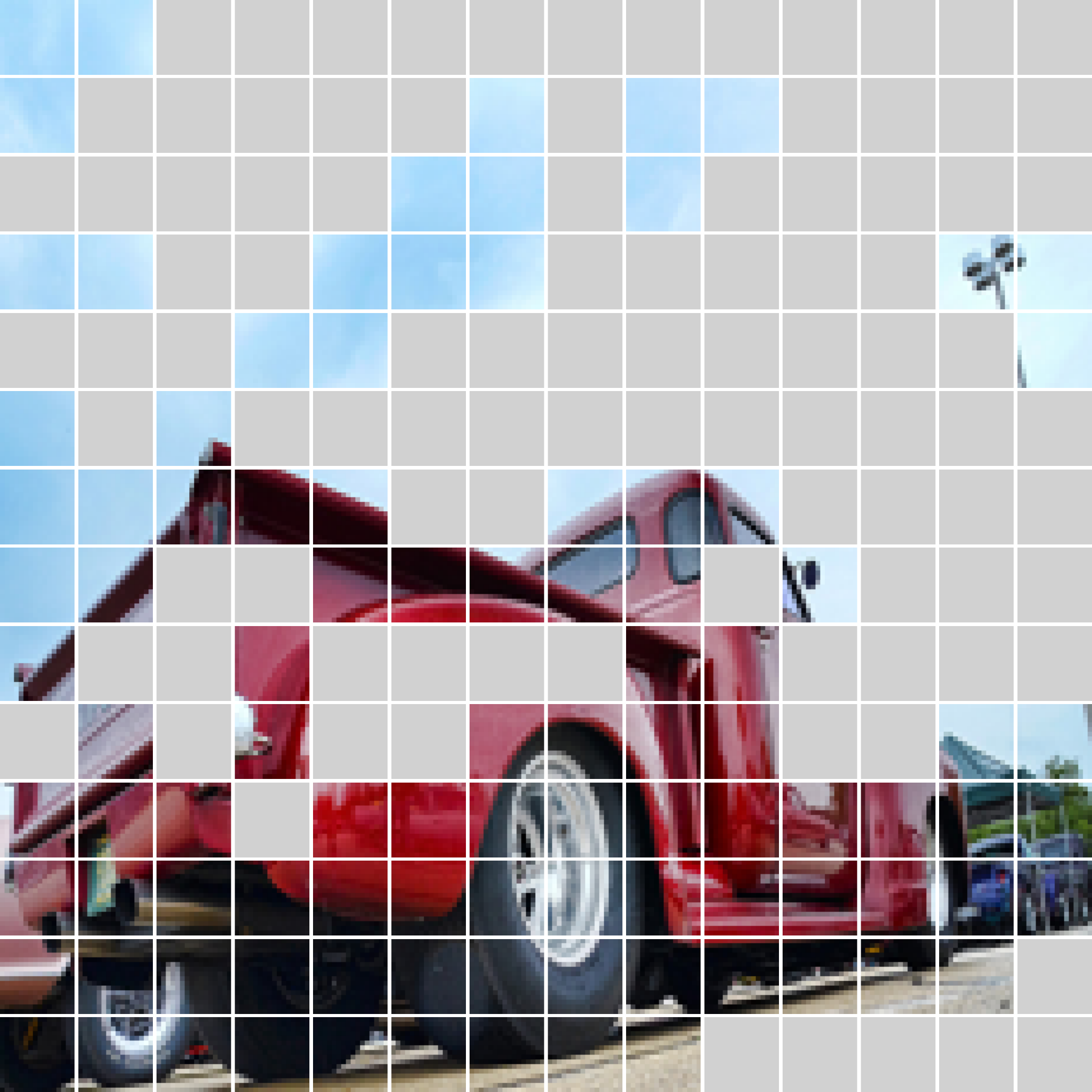}}
	\end{minipage}
    &

    \begin{minipage}[b]{0.25\columnwidth}
		\centering
		\raisebox{-.45\height}{\includegraphics[width=\linewidth]{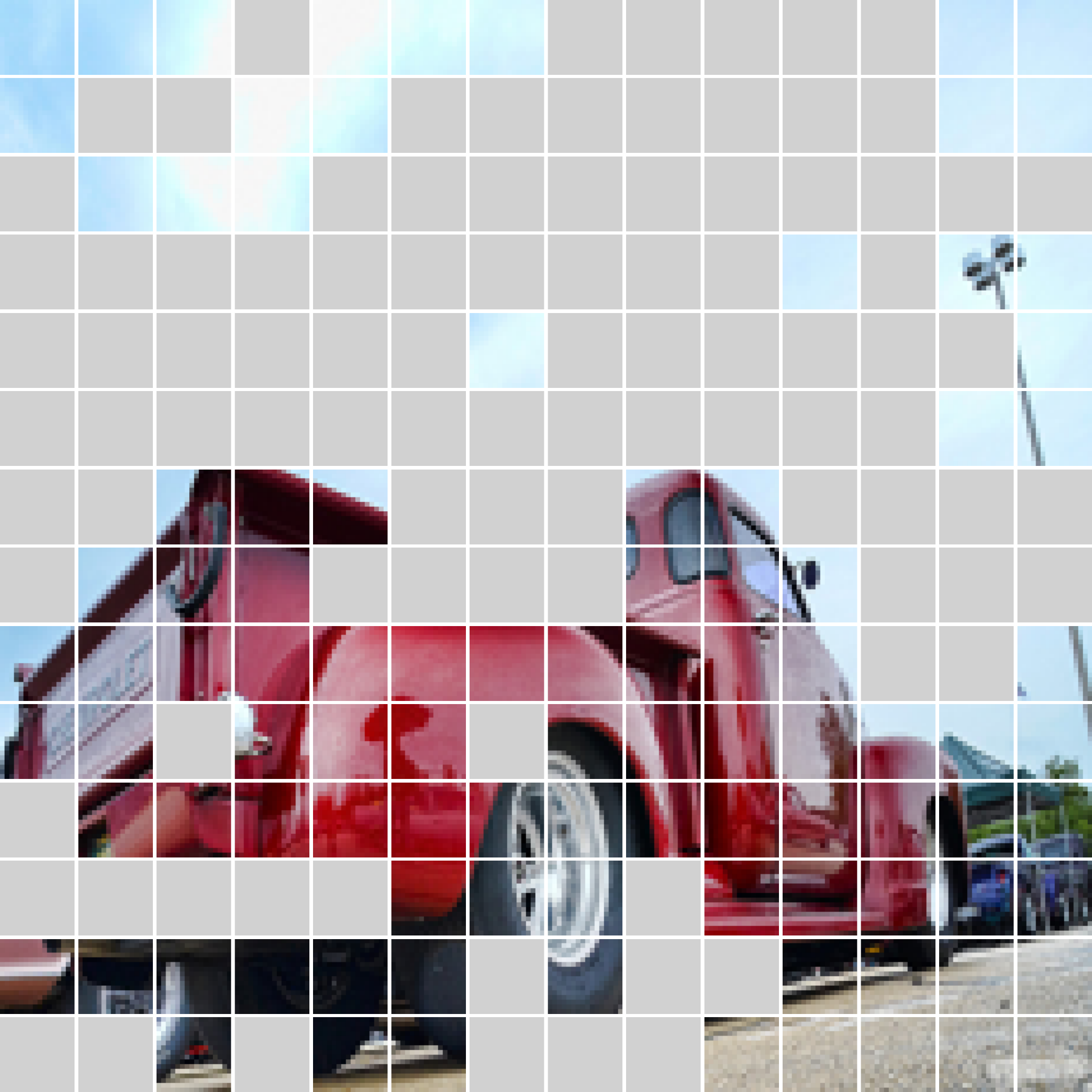}}
	\end{minipage}

    &
            \begin{minipage}[b]{0.25\columnwidth}
		\centering
		\raisebox{-.45\height}{\includegraphics[width=\linewidth]{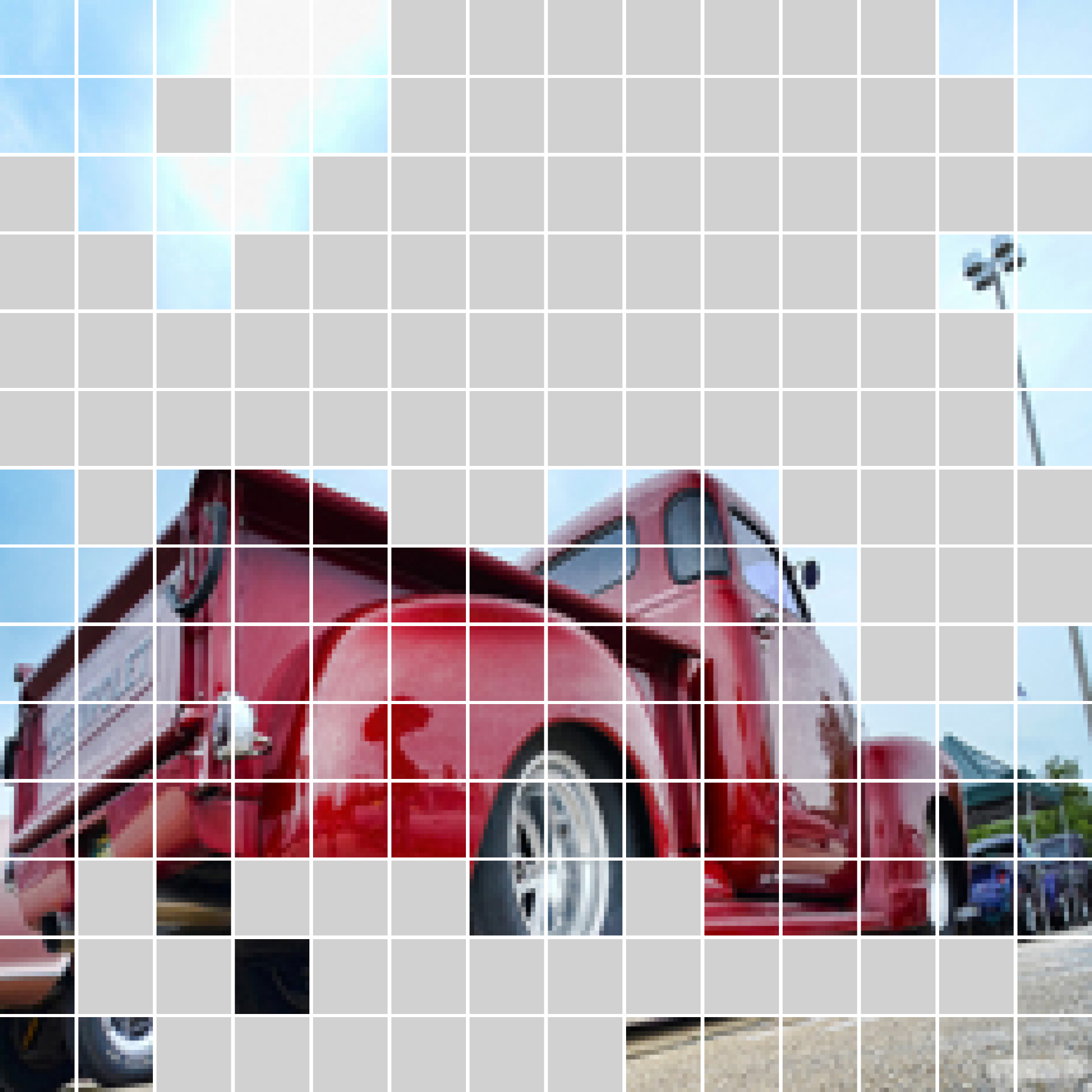}}
	\end{minipage}
   &    \begin{minipage}[b]{0.25\columnwidth}
		\centering
		\raisebox{-.45\height}{\includegraphics[width=\linewidth]{fig/teaser/teaser_cluster_masking_2.pdf}}
	\end{minipage}
    \\

    \vspace{0.018\columnwidth}
    \begin{minipage}[b]{0.25\columnwidth}
    \centering
    \raisebox{-.45\height}{\includegraphics[width=\linewidth]{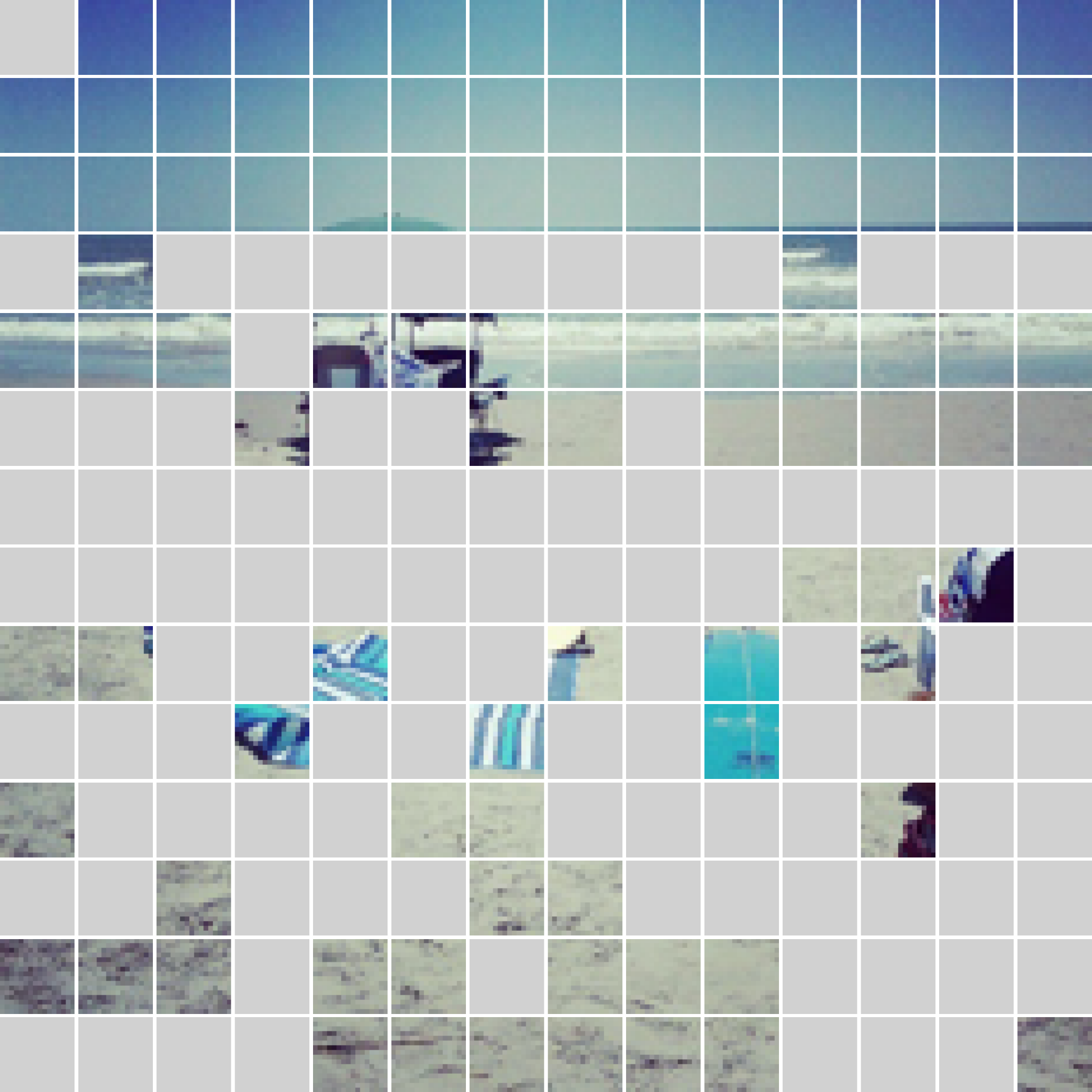}}
	\end{minipage}
    & 
    \begin{minipage}[b]{0.25\columnwidth}
		\centering
		\raisebox{-.45\height}{\includegraphics[width=\linewidth]{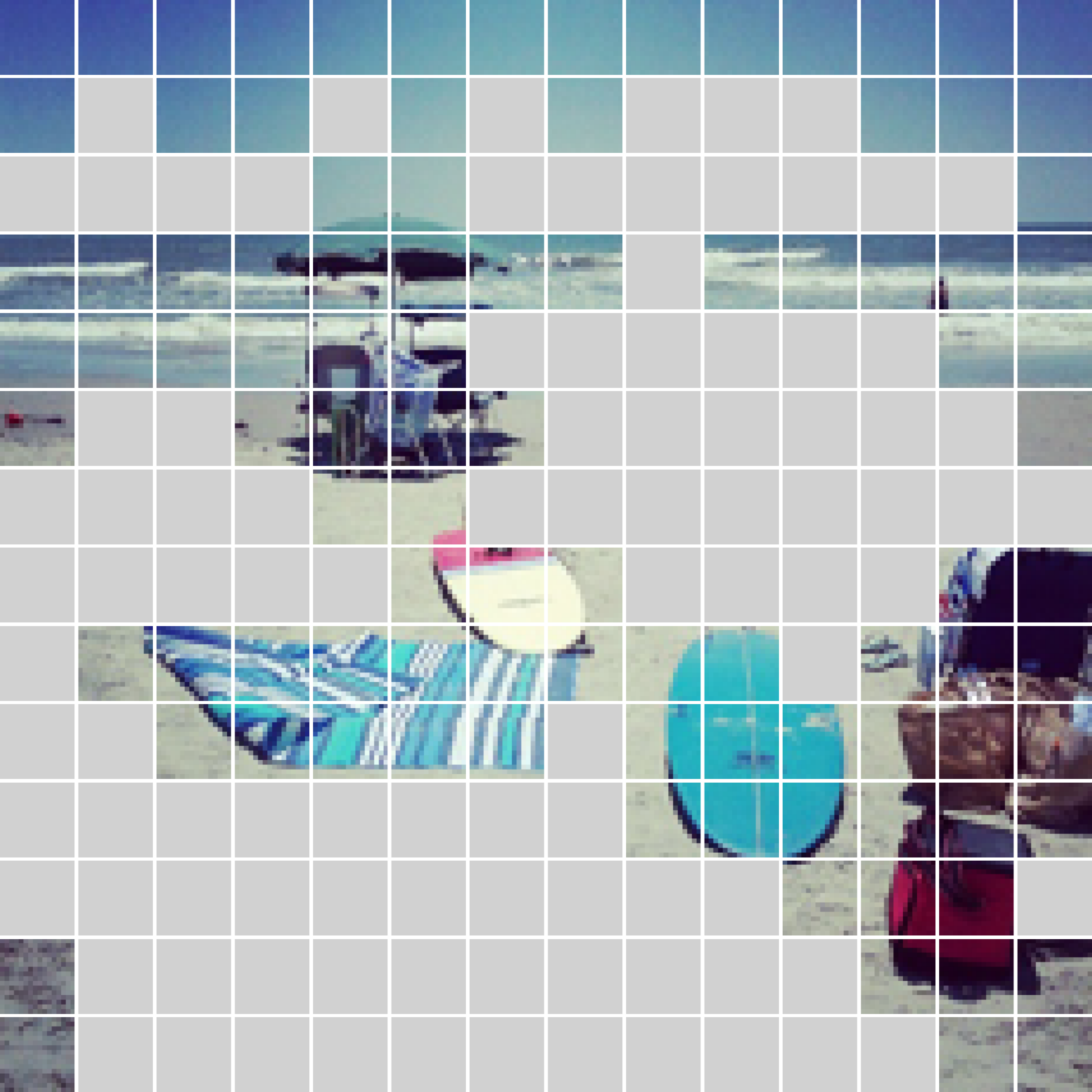}}
	\end{minipage}

    &  \begin{minipage}[b]{0.25\columnwidth}
		\centering
		\raisebox{-.45\height}{\includegraphics[width=\linewidth]{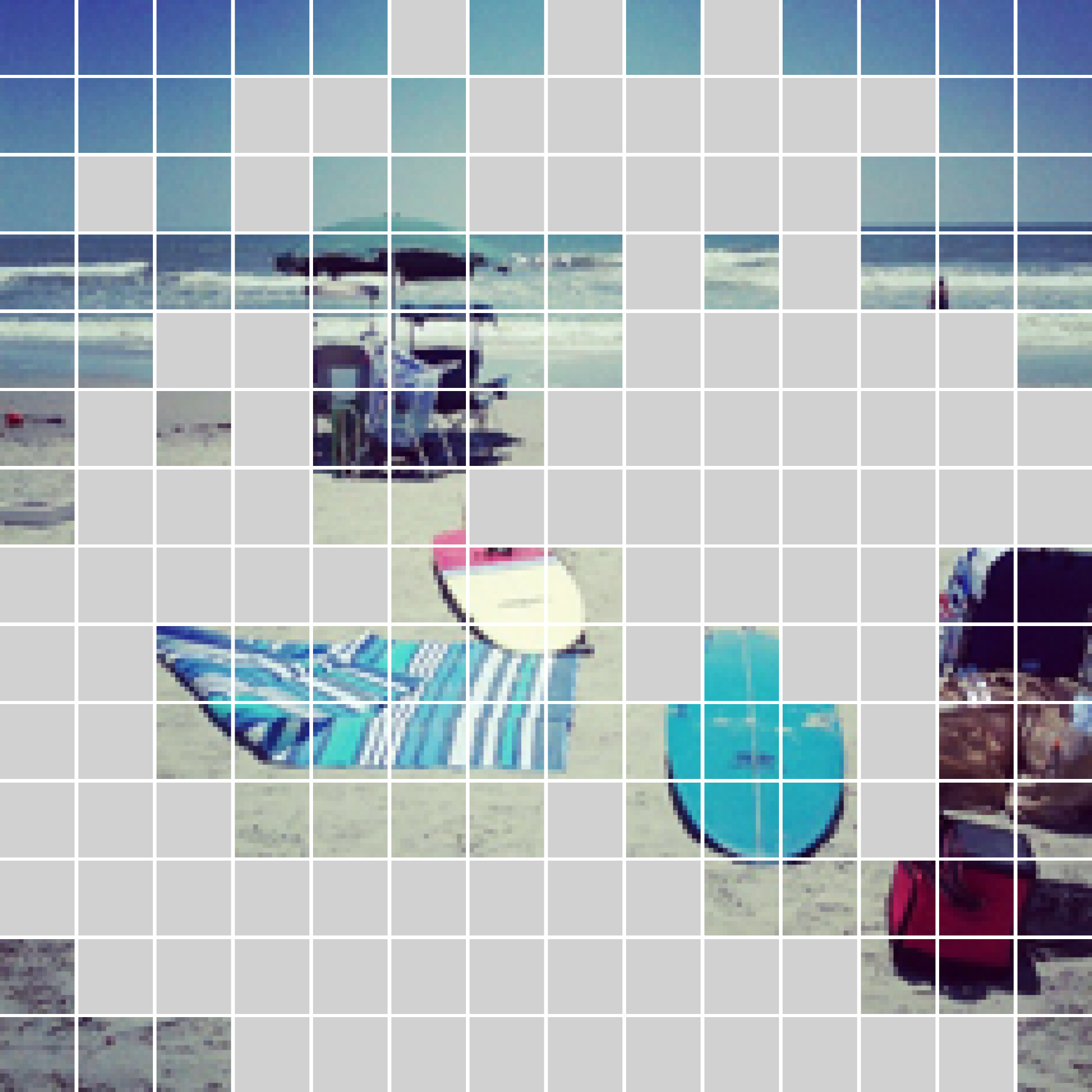}}
	\end{minipage}
  & 
   \begin{minipage}[b]{0.25\columnwidth}
		\centering
		\raisebox{-.45\height}{\includegraphics[width=\linewidth]{fig/teaser/teaser_cluster_masking_3.pdf}}
	\end{minipage}
  \end{tabular} 
  }
  \endgroup
  \caption {Visualization of attention-based masking. The images are the same from Figure \ref{fig:teaser}.}  
\label{fig:attention_mask}
\end{figure}

We extend Figure \ref{fig:teaser} with examples from the attention-guided baseline (Figure \ref{fig:attention_mask}).
In contrast to our RGB model, the behavior of the attention-based method changes during training. In early iterations, it masks randomly, while later in training it produces fairly consistent clusters that do not vary much between iterations, since the attention maps change less over time, potentially limiting the diversity of training examples.

\section{Clustering Visualization}

\begin{figure}[htbp]
  \centering
  \begingroup
  \resizebox{\linewidth}{!}{
  \includegraphics[width=\linewidth]{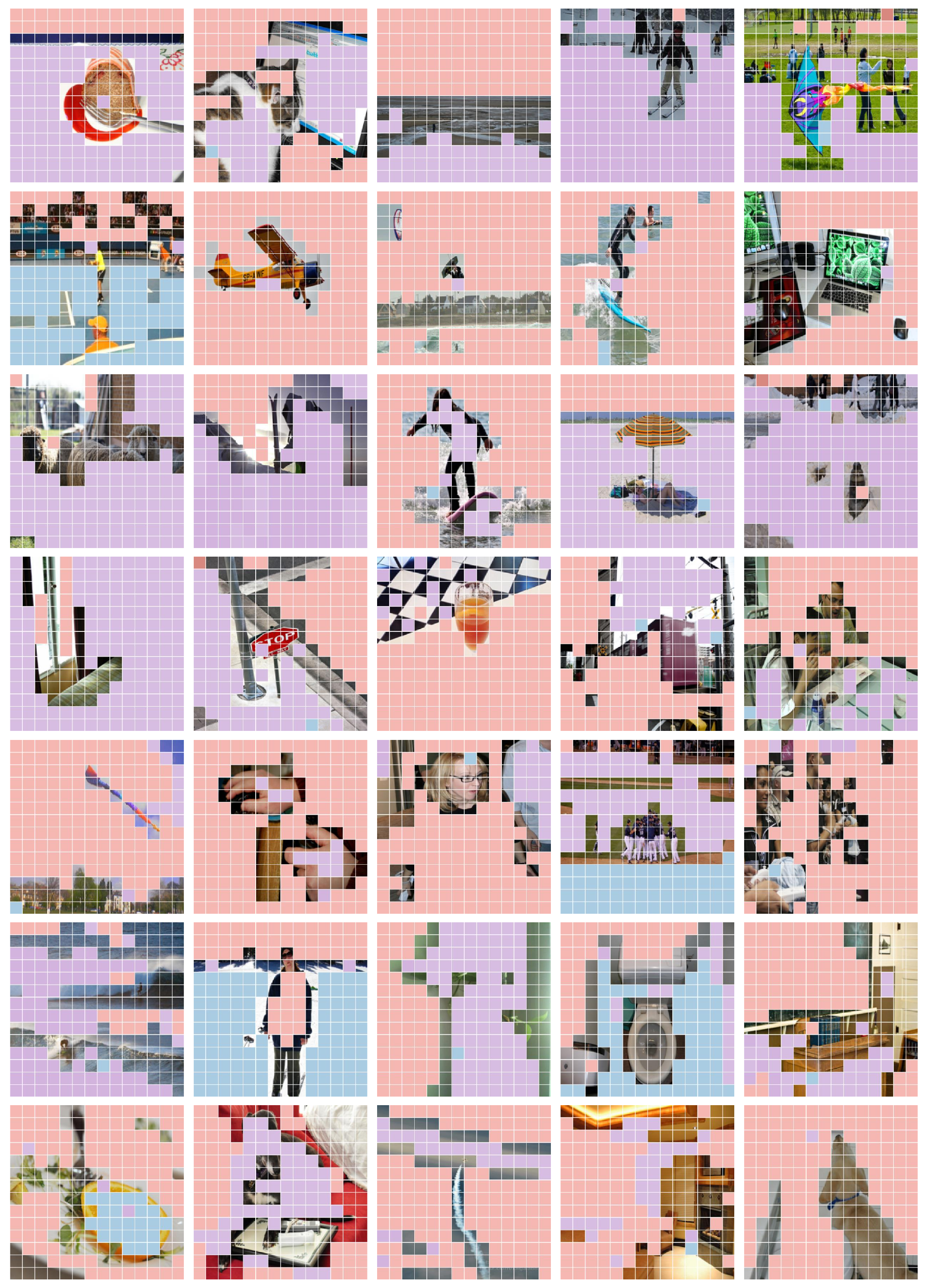}
  }
  \endgroup
  \caption {\textbf{Illustration of cluster based masks on COCO.} } 
\label{fig:visual_coco}
\end{figure}

\begin{figure}[htbp]
  \centering
  \begingroup
  \resizebox{\linewidth}{!}{
  \includegraphics[width=\linewidth]{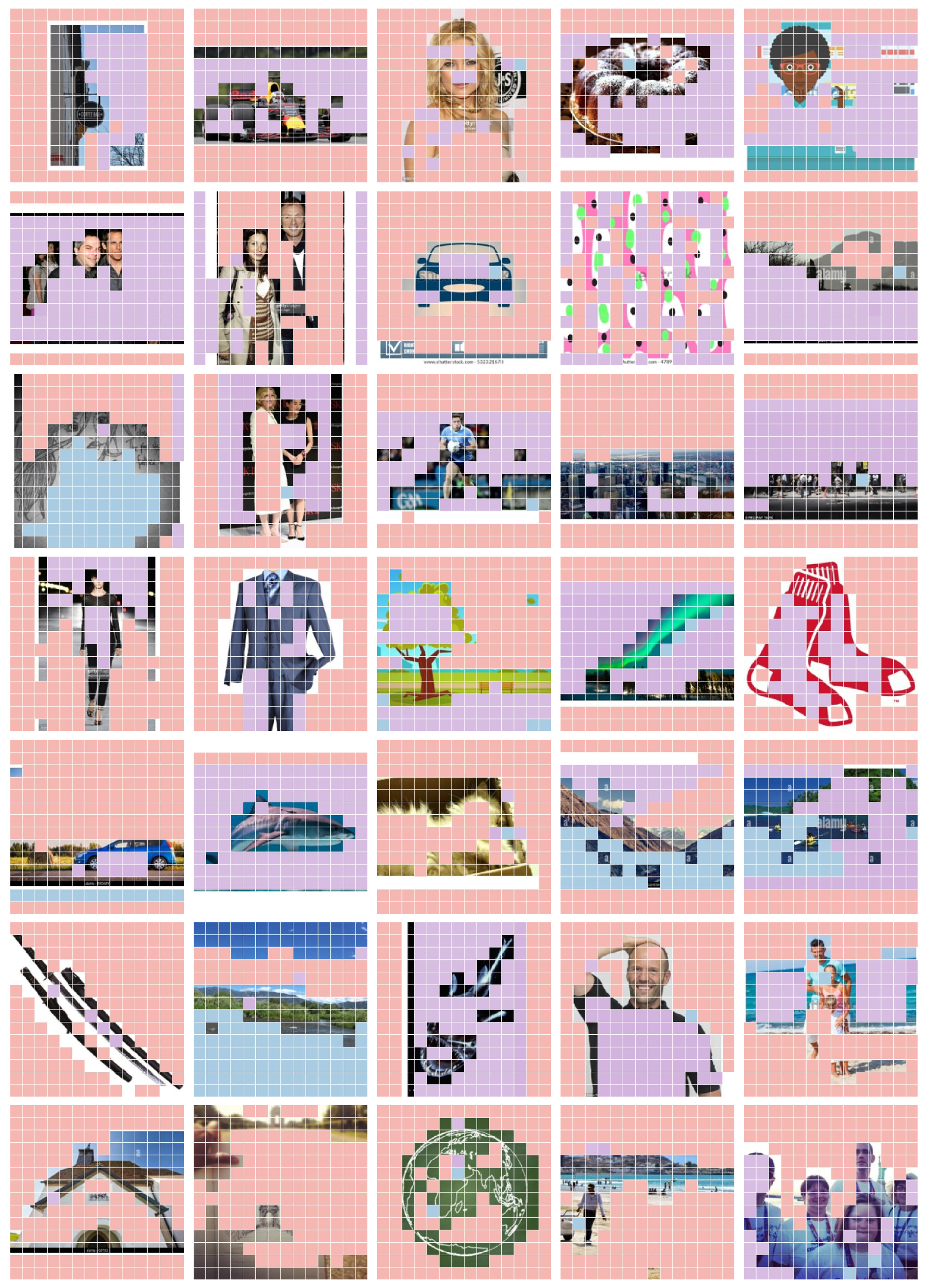}
  }
  \endgroup
  \caption {\textbf{Illustration of cluster based masks on CC3M.} } 
\label{fig:visual_cc3m}
\end{figure}

We provide more examples of our clustering masking visualization on COCO and CC3M datasets on Figure \ref{fig:visual_coco} and Figure \ref{fig:visual_cc3m} respectively. We mask out at least 50\% patches in each image.

\end{document}